%% file: main.tex
\documentclass[11pt]{article}

\usepackage[preprint]{acl}

\input{preamble}

\title{\thetitle{}}

\author{Morris Alper \\
  Carnegie Mellon University \\
  University of Miami \\\And
  Vasudha Varadarajan \\
  Carnegie Mellon University \\\And
  Moran Yanuka \\
  Tel Aviv University \\\AND
  Angelina Wang \\
  Cornell University \\\And
  Hadar Averbuch-Elor \\
  Cornell University
}

\begin{document}

\maketitle
\input{sec/00_abs}

\input{sec/01_intro}
\input{sec/02_data}

\input{sec/03_method}

\input{sec/04_exp}
\input{sec/05_rw}
\input{sec/06_conc}

\bibliography{custom}

\appendix

\input{sec/xx_app}

\end{document}

%% file: preamble.tex
\usepackage{subcaption}

\usepackage{times}
\usepackage{latexsym}

\usepackage[T1]{fontenc}

\usepackage[utf8]{inputenc}
\usepackage{makecell}
\usepackage{microtype}

\usepackage{inconsolata}

\usepackage{graphicx}

\usepackage{lipsum}
\usepackage{tabularx}
\usepackage{booktabs}
\usepackage{multirow}
\usepackage{amsmath}
\usepackage{cleveref}
\usepackage{tikz}
\usepackage{float}

\usepackage{tikz}
\usetikzlibrary{arrows.meta, positioning, calc, decorations.pathreplacing}

\newcommand{\thetitle}{\namesakes{}: Probing Identity Memorization in Text-to-Image Models}

\newcommand{\namesakes}{\textsc{Namesakes}}

\newcommand{\disp}{\ensuremath{\delta}}
\newcommand{\gtsim}{\ensuremath{s_{\mathrm{gt}}}}

\newcommand{\censim}{\ensuremath{s_{\mathrm{cen}}}}

\newcommand{\samplesize}{1,269}

\DeclareMathOperator{\tr}{tr}

\newcommand{\paperurl}{https://namesakes-web.github.io}

%% file: sec/00_abs.tex
\begin{abstract}
Text-to-image (T2I) models generate realistic likenesses of some individuals when prompted with their names, raising privacy concerns. However, distinguishing whether a generated face is memorized or fabricated currently requires ground-truth photos, access to training data, or white-box access to model internals, limiting applicability. We introduce a fully black-box behavioral probe that distinguishes between memorized and unrecognized names, while requiring no reference photos or prior knowledge of training data. To benchmark this task, we present the \namesakes{} dataset of over one thousand names and faces of public figures spanning a wide range of fame levels, along with perturbed, less famous names. Experiments on state-of-the-art T2I models show that our probe substantially predicts identity memorization and separates memorized from unrecognized names, with further insights into differences across model families.
Project page: \url{\paperurl}.
\looseness=-1
\end{abstract}

%% file: sec/01_intro.tex
\section{Introduction}
\label{sec:intro}

\input{figures/teaser}

\input{figures/fame_examples}

Text-to-image (T2I) models can now generate nearly photorealistic images from text prompts. What happens when a prompt contains a personal name? As seen in \Cref{fig:teaser}, models often reproduce celebrity faces accurately, but for unrecognized names, they often fabricate plausible faces from demographic cues. 
This difference reflects two distinct regimes. \emph{Identity memorization} refers to cases where a model reproduces a specific, real identity learned during training~\cite{gu2023memorization,hintersdorf2024finding,ma2025inversion}. We term the converse case \emph{identity fabrication}, where the model generates a plausible but non-grounded face based on demographic or semantic priors inferred from the name, related to observations of T2I demographic stereotyping~\cite{bianchi2023easily,luccioni2023stable}.

Distinguishing these regimes in vision-language settings matters for privacy, model auditing, and evaluating unlearning methods.
In particular, an individual may wish to know whether a model can generate their likeness, and auditors may need to assess memorization at scale for purposes such as regulatory compliance.
Yet existing approaches require either (a)~ground-truth (GT) photos of candidates, (b)~access to training data, or (c)~white-box access to model weights. These requirements are often unmet: State-of-the-art (SOTA) T2I models' training data is not fully disclosed, their architectures may be closed or non-standard, a comprehensive gallery of GT photos does not exist, and a user may be unwilling to upload their own photos.

In this work, we introduce a fully black-box behavioral probe designed to distinguish between memorized and fabricated generations for personal names in T2I models. This is structured without architecture-specific assumptions, and does not require access to any GT photos, model internals, or prior information about training data. In order to benchmark this method, we introduce the \namesakes{} dataset of over one thousand paired names and faces of public figures from open Wikipedia data spanning a full spectrum of fame levels, along with \textit{perturbed} %
names. We show that this benchmark effectively enables measuring our probe's effectiveness across names and models, with additional insights about the span of names recognized and identities fabricated by popular SOTA T2I models. 

We release our code\footnote{\url{https://github.com/morrisalp/namesakes}} and data\footnote{\url{https://huggingface.co/datasets/malper/namesakes}} to enable future work on privacy and interpretability of T2I models. Our data release uses non-commercial licensing, adhering to relicensing and attribution requirements, as well as stipulating ethical use requirements.

%% file: figures/teaser.tex
\newlength{\teaserimg}
\setlength{\teaserimg}{0.31\linewidth}
\newlength{\teasercol}
\setlength{\teasercol}{0.32\linewidth}
\begin{figure}[t]
  \centering
  \begin{tabular}{@{}>{\centering\arraybackslash}p{\teasercol}@{\hspace{6pt}\vrule width 1pt\hspace{6pt}}>{\centering\arraybackslash}p{\teasercol}@{}>{\centering\arraybackslash}p{\teasercol}@{}}
    \textbf{Real Face} & \multicolumn{2}{c}{\textbf{Text-to-Image Generations}} \\[4pt]
    \includegraphics[width=\teaserimg]{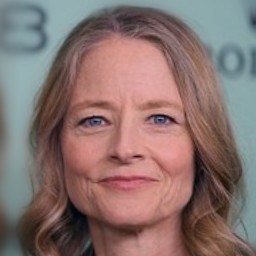} &
    \includegraphics[width=\teaserimg]{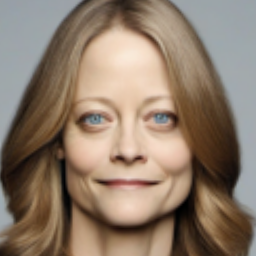} &
    \includegraphics[width=\teaserimg]{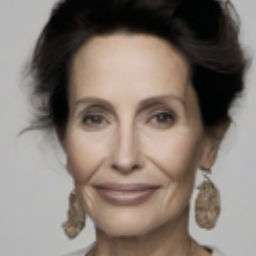} \\
    \texttt{\small{``Jodie Foster''}} & \texttt{\small{``Jodie Foster''}} & \texttt{\small{\textcolor{red}{``Jolie Fuster''}}} \\[0.25em]
    \includegraphics[width=\teaserimg]{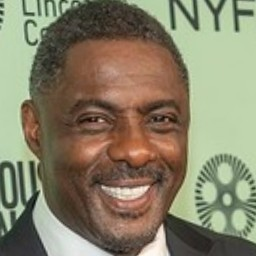} &
    \includegraphics[width=\teaserimg]{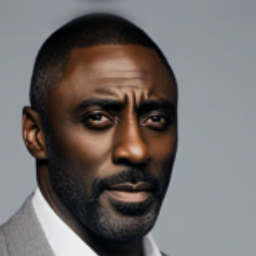} &
    \includegraphics[width=\teaserimg]{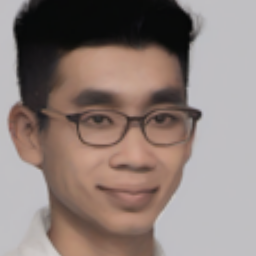} \\
    \texttt{\small{``Idris Elba''}} & \texttt{\small{``Idris Elba''}} & \texttt{\small{\textcolor{red}{``Idrus Elga''}}} \\
  \end{tabular}
  \caption{%
    T2I models memorize the faces of some individuals (left) and synthesize their likenesses when prompted with their names (middle). However, when prompted with unfamiliar names (right, \textcolor{red}{red}) these models fabricate plausible faces. Our probing method distinguishes between these
    cases without access to training data or ground-truth photos.}
  \label{fig:teaser}
\end{figure}

%% file: figures/fame_examples.tex
\begin{figure*}[h]
  \centering
  \includegraphics[width=\textwidth]{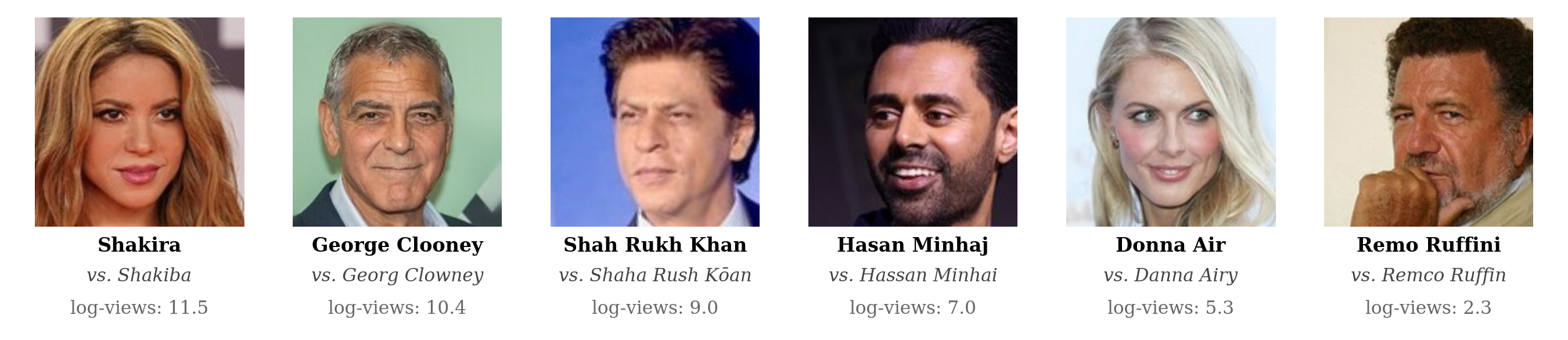}
  \vspace{-20pt}
  \caption{Samples from the \samplesize{} items in \namesakes{}. Each item consists of a public figure's name and ground-truth face (sourced from open data on Wikipedia), and fame as measured by pageview counts (log-scaled). Each real name (first line) is accompanied by a 
  perturbed 
  name (second line, after ``vs.'') designed to orthographically resemble it. Figures are chosen to span a spectrum of fame---from the average point of view of English Wikipedia users---from highly well-known (left) to relatively obscure (right).}
  \label{fig:fame_examples}
\end{figure*}

%% file: sec/02_data.tex
\input{figures/app_fame}
\input{figures/fame_scatter}

\section{\namesakes{}}
\label{sec:dataset}

We present \namesakes{}, a dataset of \samplesize{} names and paired multimodal data extracted from Wikipedia designed to serve as a benchmark for identity probing in vision-language models. We present core details here with further details provided in the appendix.

\paragraph{Dataset Contents.}
As shown in \Cref{fig:fame_examples}, each entry in \namesakes{} consists of: (1) the personal name of a public figure (from their Wikipedia entry), (2) a GT photo portraying their face, (3) an estimate of the individual's fame based on pageviews, and (4) a %
perturbed name (defined below). Names are chosen to span a wide range of fame levels, from extremely famous individuals to relatively obscure figures (still deemed public with respect to Wikipedia's inclusion criteria). Additional metadata is provided with details such as image licensing and attribution. Limitations regarding global coverage and diversity are discussed in the limitations section.

\paragraph{Dataset Construction.}
Data is sourced from English Wikipedia entries. About 190K entries have structured person information; among these, pageview counts $v$ are approximately log-normally distributed, making $f := \log(v) \sim \mathcal{N}(\cdot, \cdot)$ a natural scale. We bin $f$ and use stratified uniform sampling over $f$ (binned), filtering for entries with freely-licensed photos. This yields \samplesize{} entries including names of very famous individuals likely to be memorized by models, as well as obscure names likely to produce fabricated faces. The resulting distribution of fame levels is shown in \Cref{fig:fame_dist}.

We validate that fame is a meaningful proxy for memorization likelihood in \Cref{fig:fame_scatter}, which compares $f$ with reference similarity \gtsim{} (\Cref{sec:target}). Pearson correlation values are moderate ($r=0.36$ to $0.53$ depending on model; all $p \ll 0.001$), confirming that more famous individuals are more likely to have their identities memorized. The stratified sampling in \namesakes{} is thus designed to ensure sufficient coverage across this spectrum. Note that fame is only used to guide dataset construction; our probing method (\Cref{sec:method}) operates solely on model outputs without using external metadata.

\paragraph{Name Perturbation.}
For each name in \namesakes{}, we also construct a %
\emph{perturbed name}: 
a plausible but fictitious name that orthographically resembles the original. As seen in \Cref{fig:teaser}, paired real and %
perturbed names may differ minimally on the textual level while producing significantly different generations, as the latter is not grounded in a potentially memorized identity. %
Perturbed names are constructed by perturbing each name component while preserving its first letter (e.g., \emph{Jodie Foster} $\to$ \emph{Jolie Fuster}), by searching a large pool of names from Wikipedia for distinct names with minimum Levenshtein edit distance from the original. These new names seldom collide with existing celebrity names (see appendix); as such, they are unlikely to refer to real memorized identities, enabling our separability analysis (\Cref{sec:separability}).

We justify the use of perturbed names with consistency checks. First, we compare perturbed names 
against 100 randomly constructed names (formed by 
independently shuffling first and last names from 
the same Wikipedia pool) on SDXL-Base using the 
probing scores defined in \Cref{sec:metrics}. 
Perturbed names produce more dispersed generations 
($\delta$: $0.58$ vs.\ $0.55$, $p{=}0.008$) and 
less similar centroids ($s_\text{cen}$: $0.54$ vs.\ 
$0.56$, $p{=}0.046$), supporting their use representing the unmemorized regime. Second, we also compare the lowest-fame names (bottom 10\%) and calculate real-vs.-perturbed separability (see \Cref{sec:separability}), finding near-chance separability (AUC 0.41--0.56 across models). This further supports the use of perturbed names as proxies for unmemorized identities, as obscure real names are scored similarly to perturbed ones.

\paragraph{Train-Test Split.} \namesakes{} is designed to be used with cross-validation (CV). Our evaluation involves fitting lightweight predictive models (an OLS linear regression) to predict reference similarity $s_{gt}$ from probe scores, and a logistic regression classifier for real-vs.-%
perturbed name separability (Section \ref{sec:exp})—separately for each T2I model, since models differ substantially in their memorization behavior and generation diversity (see Table \ref{tab:models}). All metrics are calculated on held-out data folds to avoid data leakage: in particular, centroid similarity $s_{cen}$ for a test item is computed using only train-fold centroids (Appendix \ref{sec:app_stats}), ensuring that the test name's own generations do not influence its score. Cross-validation also mitigates potential issues of demographic imbalance and sensitivity to noise that could arise from a single fixed test split given our dataset size.

\paragraph{Demographic Breakdown.} As our dataset inherits subject-matter bias from the English Wikipedia, we report the demographic makeup of \namesakes{}. We use Gemini 3.1 Pro~\citep{googlegemini31} with web search grounding to annotate coarse gender and racial categories\footnote{We acknowledge our coarse annotations are necessarily reductive, and may not accurately reflect an individual's self-identification.} based on online knowledge of the public figure (for internal use only to generate representation statistics; not released), and manually validate accuracy on a 100-item subset (99\% correct).

As shown in Table~\ref{tab:demobreakdown}, \namesakes{} is majority White and men, with moderate gender balance (though lacking in non-binary representation) and larger imbalance of racial categories. Given that our benchmark's coverage reflects this composition, and known racial disparities of many computer vision methods~\cite{buolamwini2018gender, sarridis2023face}, we highlight this demographic skew as a serious limitation and encourage future work to enhance representation diversity.

\input{tables/demobreakdown}

%% file: figures/app_fame.tex
\begin{figure*}[h]
  \centering
  \includegraphics[width=\textwidth]{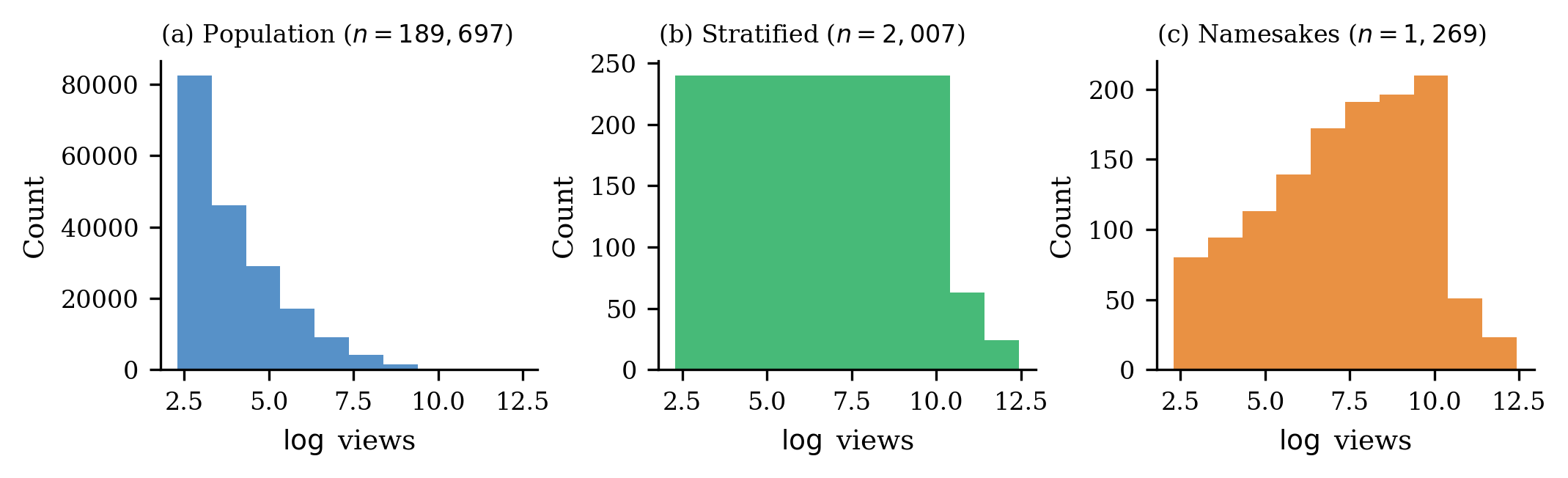}
  \vspace{-20pt}
  \caption{Distribution of fame levels (log-pageviews) in multiple stages of constructing \namesakes{}: in the initial population of entries with infoboxes (left), after stratified sampling (center), and after filtering for pages with freely-licensed infobox images (right). Overall, the final distribution covers fame levels more evenly than the initial skewed distribution. Histograms above use ten evenly-spaced bins.}
  \label{fig:fame_dist}
\end{figure*}

%% file: figures/fame_scatter.tex
\begin{figure*}[h]
  \centering
  \includegraphics[width=\textwidth]{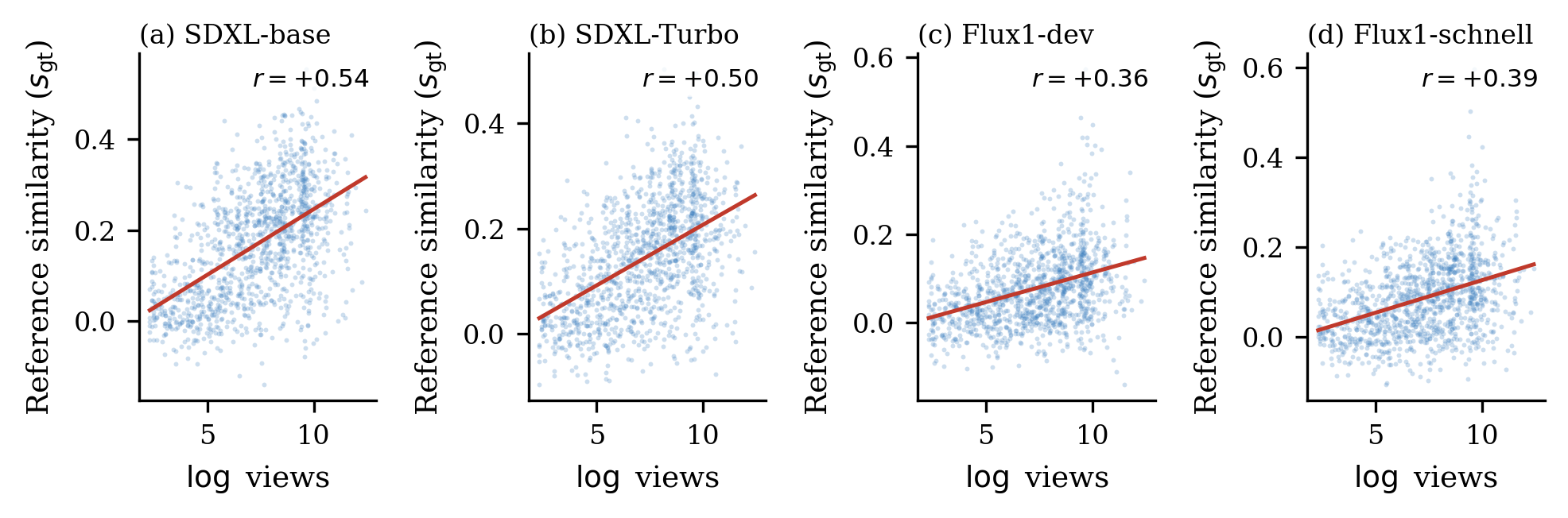}
  \vspace{-20pt}
  \caption{Plots of fame (log-pageviews) and reference similarity \gtsim{} for various T2I models under consideration, with Pearson correlation ($r$) values and best-fit lines. All settings show significant ($p \ll 0.001$) correlation, indicating that identities of more famous individuals are generally more likely to be memorized, justifying our stratified sampling in constructing \namesakes{}.}
  \label{fig:fame_scatter}
\end{figure*}

%% file: tables/demobreakdown.tex
\begin{table}[ht]
\centering
\resizebox{.9\linewidth}{!}{%
\begin{tabular}{@{}c c@{}}
\resizebox{.4\textwidth}{!}{
\begin{tabular}{l r}
\toprule
\textbf{Gender} & \textbf{Count} \\
\midrule
Men        & 757 \\
Women      & 511 \\
Non-binary  & 1 \\
\bottomrule
\end{tabular}}
&
\resizebox{.4\textwidth}{!}{\begin{tabular}{l r}
\toprule
\textbf{Race} & \textbf{Count} \\
\midrule
White        & 938 \\
Asian        & 163 \\
Black        & 73  \\
Hispanic/Latino & 45 \\
Multiracial  & 47 \\
Indigenous   & 3  \\
\bottomrule
\end{tabular}}
\end{tabular}}
\caption{Estimated Demographic Breakdown of \namesakes{} (Gender and Race)}
\label{tab:demobreakdown}
\end{table}

%% file: sec/03_method.tex
\input{figures/probe}

\section{Method}
\label{sec:method}

Given a T2I model and name, our reference-free probing method aims to predict whether the model has memorized the name's associated identity, i.e., whether generations will resemble the individual's likeness. We formalize this by defining a ground-truth measure of memorization (reference similarity) and then designing behavioral probes that predict it without access to any reference photos. We emphasize that reference photos are not used in the probe itself; reference similarity is only used in our experiments for a post-hoc evaluation of probe effectiveness.

\subsection{Preliminaries}
\label{sec:prelim}

To probe for knowledge of identities, our method requires a similarity metric: given facial images $I_i$ and $I_j$, the similarity score should be high when the same person is depicted in both images and low otherwise. We implement this using a facial recognition encoder which assigns embeddings to these images and calculates similarity via their cosine similarity. This serves as a proxy for identity similarity, while inheriting the limitations of current facial recognition methods as discussed in \Cref{sec:limitations}.

Given a name with GT reference image $G$ and a given T2I model we generate $k$ images using a fixed textual prompt template and input noise seeds. The respective L2-normalized face embeddings are denoted by $e_{\mathrm{gt}}$ and $e_1, \cdots, e_k$ (generations). We also denote the centroid of generation embeddings as $\bar{e} := \frac{1}{k}\sum_i e_i$, L2-normalized to $\hat{e} := \frac{1}{\|\bar{e}\|}\bar{e}$.

\subsection{Reference Similarity (\gtsim)}
\label{sec:target}

We define reference similarity as the mean similarity between the ground-truth (GT) and generated facial images:
\begin{equation}
  \gtsim := \frac{1}{k}\sum_{i=1}^{k}e_i \cdot e_{\mathrm{gt}} = \bar{e} \cdot e_{\mathrm{gt}}.
\end{equation}
Reference similarity directly operationalizes identity memorization: when \gtsim{} is high, the model reproduces the real person's likeness (identity memorization); when low, it generates a generic face unrelated to the individual (identity fabrication). Our goal is to predict this quantity \emph{without access to the ground-truth appearance}, enabling black-box memorization detection. Accordingly, the subsequent probing scores do not have access to $e_{\mathrm{gt}}$.

\subsection{Probing Scores}
\label{sec:metrics}

We define two complementary scores, \disp{} and \censim{}, designed to predict reference similarity, and thereby identity memorization, without access to ground-truth photos, illustrated in \Cref{fig:probes}. These are both entirely black-box and reference-free; in addition, \disp{} only uses a single name as input, while \censim{} also requires comparisons to generations for other names in \namesakes{}. We will later (\Cref{sec:exp}) show that \disp{} can be used alone with substantial performance; alternatively, \censim{} may be added for an additional boost.

\paragraph{Dispersion (\disp).}
This quantity measures inter-generation consistency, i.e., to what degree multiple generations of a name depict the same face. This is defined as the mean squared distance 
\begin{equation}
  \disp := \frac{1}{k}\sum_{i=1}^{k}\left\| e_i - \bar{e} \right\|^2 = \tr(\Sigma),
\end{equation}
where $\Sigma$ is the sample covariance matrix of $(e_1, \cdots, e_k)$. Intuitively, memorized identities should result in more similar generated faces, negatively correlating \disp{} with \gtsim{}.

\paragraph{Centroid similarity (\censim).} Let $\hat{\mathcal{N}}$ denote the set of all L2-normalized centroids of other names in \namesakes{}, and define
\begin{equation}
  \censim := \max_{\hat{e}' \in \hat{\mathcal{N}}} \hat{e} \cdot \hat{e}'.
\end{equation}
Intuitively, unfamiliar names are expected to produce generic outputs that closely resemble those of other names, while a memorized identity should only be produced by its associated name. Hence, \censim{} is expected to negatively correlate with \gtsim{}, consistent with findings that T2I models converge to default images for unknown prompts~\citep{defaultimages}.

%% file: figures/probe.tex
\newlength{\probeimg}
\setlength{\probeimg}{0.075\textwidth}
\newlength{\probeimgsm}
\setlength{\probeimgsm}{0.065\textwidth}

\begin{figure*}[t]
  \centering
  \resizebox{\linewidth}{!}{
  \begin{tikzpicture}[
      img/.style={inner sep=0pt, outer sep=0pt},
      lbl/.style={font=\small\ttfamily, align=center},
      annot/.style={font=\small, text=black!70},
      sectitle/.style={font=\bfseries\large},
      reglbl/.style={font=\small\bfseries, text=black!70},
      centnode/.style={draw, circle, minimum size=18pt, inner sep=0pt,
                       font=\normalsize, fill=black!5},
      simlink/.style={<->, black!35, shorten >=2pt, shorten <=2pt},
      >=Stealth,
    ]

    \def\ytitle{0}
    \def\yrow{-0.5}

    \node[sectitle] at (2.2, \ytitle) {Dispersion ($\delta$)};

    \def\xmleft{0}
    \node[reglbl] at (\xmleft+0.6, \yrow-0.2) {Memorized ($\downarrow\delta$)};

    \def\mtx{16pt}
    \def\mty{16pt}
    \node[img] (m1) at (\xmleft - 0.3, \yrow - 1.3)
      {\includegraphics[width=\probeimg]{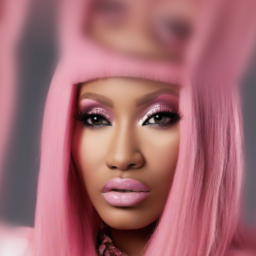}};
    \node[img, right=\mtx of m1] (m2)
      {\includegraphics[width=\probeimg]{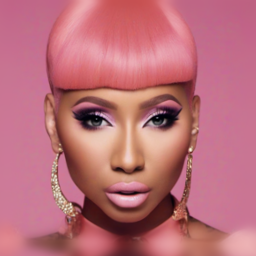}};
    \node[img, below=\mty of m1] (m3)
      {\includegraphics[width=\probeimg]{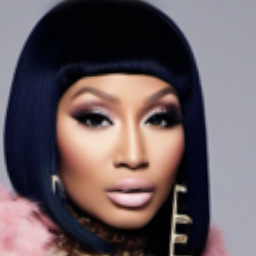}};
    \node[img, below=\mty of m2] (m4)
      {\includegraphics[width=\probeimg]{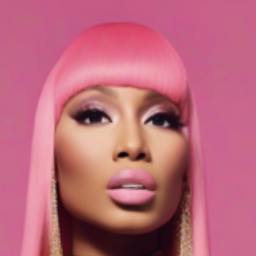}};

    \draw[simlink] (m1.east) -- (m2.west);
    \draw[simlink] (m3.east) -- (m4.west);
    \draw[simlink] (m1.south) -- (m3.north);
    \draw[simlink] (m2.south) -- (m4.north);
    \draw[simlink] (m1.south east) -- (m4.north west);
    \draw[simlink] (m2.south west) -- (m3.north east);

    \node[lbl, below=5pt of $(m3.south)!0.5!(m4.south)$] (mname) {``Nicki Minaj''};

    \def\xhright{4.3}
    \node[reglbl] at (\xhright, \yrow-0.2) {Fabricated ($\uparrow\delta$)};

    \def\htx{32pt}
    \def\hty{28pt}
    \node[img] (h1) at (\xhright - 0.9, \yrow - 1.3)
      {\includegraphics[width=\probeimg]{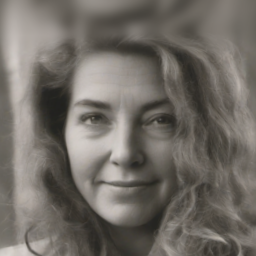}};
    \node[img, right=\htx of h1] (h2)
      {\includegraphics[width=\probeimg]{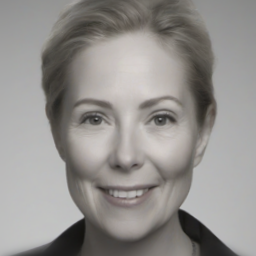}};
    \node[img, below=\hty of h1] (h3)
      {\includegraphics[width=\probeimg]{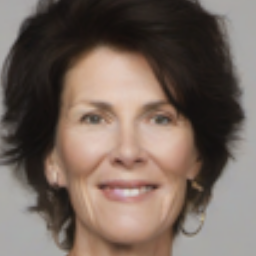}};
    \node[img, below=\hty of h2] (h4)
      {\includegraphics[width=\probeimg]{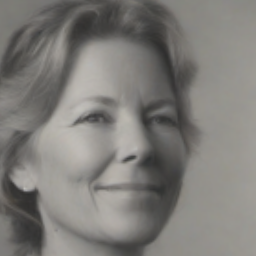}};

    \draw[simlink] (h1.east) -- (h2.west);
    \draw[simlink] (h3.east) -- (h4.west);
    \draw[simlink] (h1.south) -- (h3.north);
    \draw[simlink] (h2.south) -- (h4.north);
    \draw[simlink] (h1.south east) -- (h4.north west);
    \draw[simlink] (h2.south west) -- (h3.north east);

    \node[lbl, below=5pt of $(h3.south)!0.5!(h4.south)$] (hname) {``Kim Stolz''};

    \def\xdiv{7.0}
    \draw[black!20, line width=0.4pt]
      (\xdiv, \ytitle + 0.3) -- (\xdiv, \yrow - 4.0);

    \def\xright{11.4}
    \node[sectitle] at (\xright, \ytitle) {Centroid similarity ($s_{\text{cen}}$)};

    \node[reglbl] at (\xright, \yrow-0.2) {Memorized ($\downarrow s_{\text{cen}}$)};

    \node[img] (a1) at (\xright - 3.2, \yrow - 1.1)
      {\includegraphics[width=\probeimgsm]{media/probes/sdxl_nicki_minaj_0.png}};
    \node[img, right=2pt of a1] (a2)
      {\includegraphics[width=\probeimgsm]{media/probes/sdxl_nicki_minaj_1.png}};

    \node[right=4pt of a2, font=\small] (arrA) {$\to$};
    \node[centnode, right=2pt of arrA] (cenA) {$\hat{e}$};
    \node[right=5pt of cenA, font=\large, text=red!60!black] (neqsym) {$\neq$};
    \node[centnode, right=5pt of neqsym] (cenB) {$\hat{e}'$};
    \node[right=2pt of cenB, font=\small] (arrB) {$\leftarrow$};

    \node[img, right=4pt of arrB] (b1)
      {\includegraphics[width=\probeimgsm]{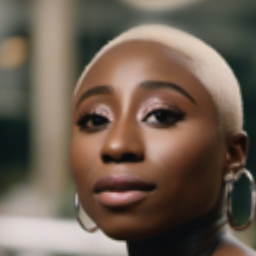}};
    \node[img, right=2pt of b1] (b2)
      {\includegraphics[width=\probeimgsm]{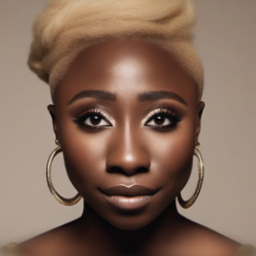}};

    \node[lbl, below=4pt of $(a1.south)!0.5!(a2.south)$] {``Nicki Minaj''};
    \node[lbl, below=4pt of $(b1.south)!0.5!(b2.south)$] {``Cynthia Erivo''};

    \def\yhallr{-2.8}
    \node[reglbl] at (\xright, \yhallr-0.2) {Fabricated ($\uparrow s_{\text{cen}}$)};

    \node[img] (c1) at (\xright - 3.2, \yhallr - 1.1)
      {\includegraphics[width=\probeimgsm]{media/probes/sdxl_kim_stolz_0.png}};
    \node[img, right=2pt of c1] (c2)
      {\includegraphics[width=\probeimgsm]{media/probes/sdxl_kim_stolz_1.png}};

    \node[right=4pt of c2, font=\small] (arrC) {$\to$};
    \node[centnode, right=2pt of arrC] (cenC) {$\hat{e}$};
    \node[right=5pt of cenC, font=\large, text=blue!60!black] (apxsym) {$\approx$};
    \node[centnode, right=5pt of apxsym] (cenD) {$\hat{e}'$};
    \node[right=2pt of cenD, font=\small] (arrD) {$\leftarrow$};

    \node[img, right=4pt of arrD] (d1)
      {\includegraphics[width=\probeimgsm]{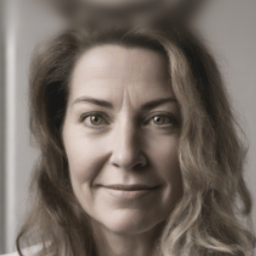}};
    \node[img, right=2pt of d1] (d2)
      {\includegraphics[width=\probeimgsm]{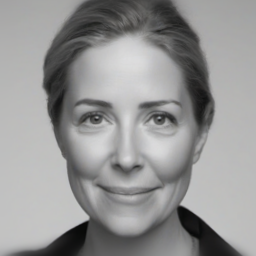}};

    \node[lbl, below=4pt of $(c1.south)!0.5!(c2.south)$]
      {``Kim Stolz''};
    \node[lbl, below=4pt of $(d1.south)!0.5!(d2.south)$]
      {``Lisa Ferraday''};

  \end{tikzpicture}}

  \caption{%
    The two probes comprising our method, shown on memorized
    vs.\ fabricated names (SDXL-Base).
    \textbf{Left --- Dispersion ($\delta$):}
    Four generations of a memorized name form a tight cluster
    (low~$\delta$), while generations of a fabricated name
    spread apart (high~$\delta$).
    Lines show pairwise comparisons; spacing reflects embedding distance.
    This comparison is \emph{within} a single name's generations.
    \textbf{Right --- Centroid similarity ($s_{\text{cen}}$):}
    Each name's generations are summarized by an embedding
    centroid~$\hat{e}$.
    Memorized names have distinctive centroids
    ($\hat{e}\neq\hat{e}'$), while fabricated names with
    similar demographic associations converge to nearly
    identical centroids ($\hat{e}\approx\hat{e}'$).
    This comparison is \emph{between} names.
  }
  \label{fig:probes}
\end{figure*}

%% file: sec/04_exp.tex
\input{tables/probe}

\section{Experiments}
\label{sec:exp}
\paragraph{Experimental Setup}
We evaluate four T2I models covering recent text-conditioned denoising diffusion models, in both base and distilled variants: Stable Diffusion XL-base (SDXL)~\citep{podellsdxl}, SDXL-Turbo~\citep{sauer2024adversarial}, Flux1-Dev, and Flux1-Schnell~\citep{flux2024}. We generate $k\!=\!4$ images per name and model for all real and perturbed
names in the \namesakes{} benchmark of $n\!=\!\samplesize{}$ items.
We ablate the prompt template, number of generated images, and the name pool size used to calculate $\censim$ in the appendix.
Generated faces undergo standard preprocessing including facial alignment, and are then embedded with ArcFace~\citep{deng2019arcface}.
For linear probes combining $\delta$ and $s_{cen}$, coefficients of each are fit per model, as models differ in their relative contributions; this fitting is negligible in cost relative to image generation.
All metrics are calculated on held-out data using 5-fold CV. Additional ablations on the choice of face embedding model are shown in the appendix.

\paragraph{Predicting Reference Similarity}
We test the effectiveness of our probe scores to predict reference similarity via ordinary least squares (OLS) linear regression; as described in \Cref{sec:method}, this aims to measure the degree of memorization as a continuous scalar value. Results for all models are shown in \Cref{tab:models} (left columns). The two-probe model achieves $R^2$ values above Cohen’s large-effect benchmark~\cite{cohen2013statistical}, with $0.58$ for SDXL-Base and between $0.33$--$0.44$ for the remaining models, indicating that our reference-free probes capture a substantial portion of the memorization signal that would otherwise require ground-truth photos. We note that reference similarity is itself a noisy proxy for true identity memorization (limited by face embedding accuracy and generation variance), which places an inherent ceiling on achievable $R^2$.

The variation in probe effectiveness between models is consistent with known differences between them: distilled models (SDXL-Turbo, Flux1-Schnell) and model families optimized for aesthetic quality (e.g., Flux) are known to have reduced diversity in generations~\citep{gandikota2026distilling,adamkiewicz2026pretty}, potentially weakening the effectiveness of our dispersion probe. In addition, Flux models have been observed to underperform on generating celebrities, leading to informal community speculation on whether they were trained with methods such as image recaptioning or targeted redaction of personal names~\citep{github,reddit}.

We also see that both probes contribute to overall performance, although for SDXL-Base dispersion dominates. This is supported by observing that \disp{} and \censim{} have Pearson correlation $0.25$ for SDXL-Base and are nearly orthogonal (between $0.04$ and $0.07$) for other models. We hypothesize that in cases where generations are less diverse, \disp{} provides much weaker signal, increasing the relative utility of \censim{}.

\paragraph{Real-vs.-Perturbed %
Separability}
\label{sec:separability}
Beyond continuous prediction of memorization magnitude as a scalar, we further address the practically relevant question of whether our probe can categorically separate memorized from unrecognized names. To this end, we apply our probe to the real and perturbed 
names in \namesakes{} (\Cref{sec:dataset}), fitting a logistic regression classifier to real vs. perturbed %
names in each CV train split and evaluating AUC and binary accuracy on held-out data.
Note that the two classes are balanced by design, as each real name has a corresponding perturbed name. Results are shown in \Cref{tab:models} (right column), finding that the probe generally separates real from perturbed names across all models. While SDXL-Base again shows the strongest performance ($\text{AUC}=0.86, \text{Acc}=0.79$), separability remains moderate ($\text{AUC} > 0.77, \text{Acc} > 0.72$) for models where continuous prediction of \gtsim{} is weaker, suggesting that this probe also serves as a binary predictor of identity.
This aggregate score reflects the fame distribution of \namesakes{} \emph{by design}: low-fame names are rarely memorized, so a well-calibrated probe should treat them as indistinguishable from their  perturbations, the intended behavior, as the probe targets memorization rather than name familiarity. Restricting to high-fame individuals (the regime where memorization is plausible) confirms this, yielding substantially stronger results (e.g., AUC of $0.95$ and ${>}90\%$ binary accuracy for SDXL-Base; see \Cref{tab:models}).

\input{figures/qual_gallery}
\input{figures/qual_cross_model}

\paragraph{Demographic Disaggregation}
In \Cref{sec:app_demog_disagg}, we disaggregate the results in \Cref{tab:models} by demographic annotations, finding that $R^2$ (predicting reference similarity) is lower for faces labeled Black or woman, while these disparities go away or even reverse for AUC (real-vs.-perturbed separability). Possible explanations for this divergence could include demographic differences in name perturbations or bias in the reference similarity target, which also relies on face embeddings~\citep{buolamwini2018gender,yucer2024racial}. Any uses of our framework should be sure to account for possible quality-of-service disparities.

\paragraph{Qualitative results}
\Cref{fig:qual_gallery} shows qualitative examples of generations (SDXL-Base) conditioned on names, illustrating different regimes for memorized and unknown names--in the former case, generations are mutually consistent and resemble the GT face, while in the latter case, they reflect more dispersed default images unrelated to the GT face, reflected in \gtsim{} and probe values. Similar results for all models tested, with more names, are shown in the appendix.

In \Cref{fig:qual_cross_model}, we compare results between T2I models for a celebrity name, illustrating the patterns seen in our quantitative results. SDXL-Based models reproduce the individual's identity more faithfully ($\gtsim{} > 0.37$), while Flux-based models fabricate generic faces that do not match the ground-truth face ($\gtsim{} < 0.07$). This matches our findings that SDXL models (particularly SDXL-Base) exhibit more identity memorization. We also see that probe values inversely correlate with \gtsim{}, as expected.

In the appendix, we also introduce an interpretability technique for visualizing overall visual associations with unfamiliar names.

\input{figures/demog_bar}

\paragraph{Human Validation}

To validate our automatic metrics (reference similarity \gtsim{} and probes) against human perception, we conduct a face similarity study on Amazon Mechanical Turk. We evaluate 100 celebrities from \namesakes{}, stratified to be demographically balanced as well as covering the spectrum of fame levels, asking annotators to indicate the similarity between the GT photo and SDXL-Base generation for a given name on a 5-point Mean Opinion Score (MOS) scale. We collect 9 annotations per item (after filtering for control tests). Further annotation task details are provided in \Cref{sec:survey-interface}.

In general, we find a substantial correlation between \gtsim{} and human similarity judgments (Pearson $r\!=\!0.56$, considered a large effect by Cohen's conventions \cite{cohen2013statistical}), with large differences between racial (Asian $r\!=\!0.36$; Black $r\!=\!0.44$; Hispanic/Latino $r\!=\!0.72$; White $r\!=\!0.74$) and gender (man $r\!=\!0.65$; woman $r\!=\!0.47$) categories. The correlation between our OLS scores (\Cref{sec:exp}) and human similarity judgments shows the same general pattern shown in Figure~\ref{fig:ols_vs_avgsim} (i.e., lower on faces of Black and Asian individuals, and women). While our automatic metrics still have a moderate correlation with similarity judgments across categories, they inherit unequal performance from the facial recognition models used for similarity calculations~\citep{buolamwini2018gender,yucer2024racial}.
This highlights the importance of mitigating bias in these models for downstream methods such as ours. Users should account for this context in interpreting results of our system, as we discuss in our ethical considerations section.

%% file: tables/probe.tex
\begin{table*}[!ht]
  \centering
  \resizebox{\linewidth}{!}{
  \begin{tabular}{lccccccc}
  \toprule
  & \multicolumn{3}{c}{Predicting Reference Similarity ($R^2$)} & \multicolumn{4}{c}{Real vs.\ Perturbed Names %
  Separability} \\
  \cmidrule(lr){2-4} \cmidrule(lr){5-8}
  Model & \disp{} only & \censim{} only & Both & AUC\textsubscript{all} & Acc\textsubscript{all} & AUC\textsubscript{high} & Acc\textsubscript{high} \\
  \midrule
    SDXL-Base & 0.547 {\scriptsize $\pm$0.031} & 0.128 {\scriptsize $\pm$0.030} & 0.581 {\scriptsize $\pm$0.040} & 0.859 {\scriptsize $\pm$0.015} & 0.791 {\scriptsize $\pm$0.014} & 0.947 {\scriptsize $\pm$0.011} & 0.902 {\scriptsize $\pm$0.014} \\
    SDXL-Turbo & 0.288 {\scriptsize $\pm$0.050} & 0.171 {\scriptsize $\pm$0.052} & 0.438 {\scriptsize $\pm$0.044} & 0.772 {\scriptsize $\pm$0.036} & 0.723 {\scriptsize $\pm$0.029} & 0.835 {\scriptsize $\pm$0.024} & 0.777 {\scriptsize $\pm$0.019} \\
    Flux1-Dev & 0.218 {\scriptsize $\pm$0.054} & 0.150 {\scriptsize $\pm$0.007} & 0.349 {\scriptsize $\pm$0.060} & 0.781 {\scriptsize $\pm$0.033} & 0.750 {\scriptsize $\pm$0.018} & 0.888 {\scriptsize $\pm$0.021} & 0.832 {\scriptsize $\pm$0.021} \\
    Flux1-Schnell & 0.137 {\scriptsize $\pm$0.078} & 0.187 {\scriptsize $\pm$0.036} & 0.325 {\scriptsize $\pm$0.056} & 0.785 {\scriptsize $\pm$0.016} & 0.748 {\scriptsize $\pm$0.011} & 0.844 {\scriptsize $\pm$0.018} & 0.792 {\scriptsize $\pm$0.015} \\
  \bottomrule
  \end{tabular}}
  \caption{
    Per-model probe results on $n\!=\!1{,}269$ items in \namesakes{} (excluding failed face alignment; see appendix).
    Left: $R^2$ from OLS linear regression predicting reference similarity \gtsim{}, using each probe alone or both.
    Right: Real vs.\ perturbed %
    separability using logistic regression on both probe features (\textit{all}: all names; \textit{high}: high-fame names with log-pageviews above median).
    Full fame stratification in appendix.
    All results use cross-validation (mean\,$\pm$\,std across folds).
  }
  \label{tab:models}
\end{table*}

%% file: figures/qual_gallery.tex
\begin{figure*}[!ht]
  \centering
  \includegraphics[width=\textwidth]{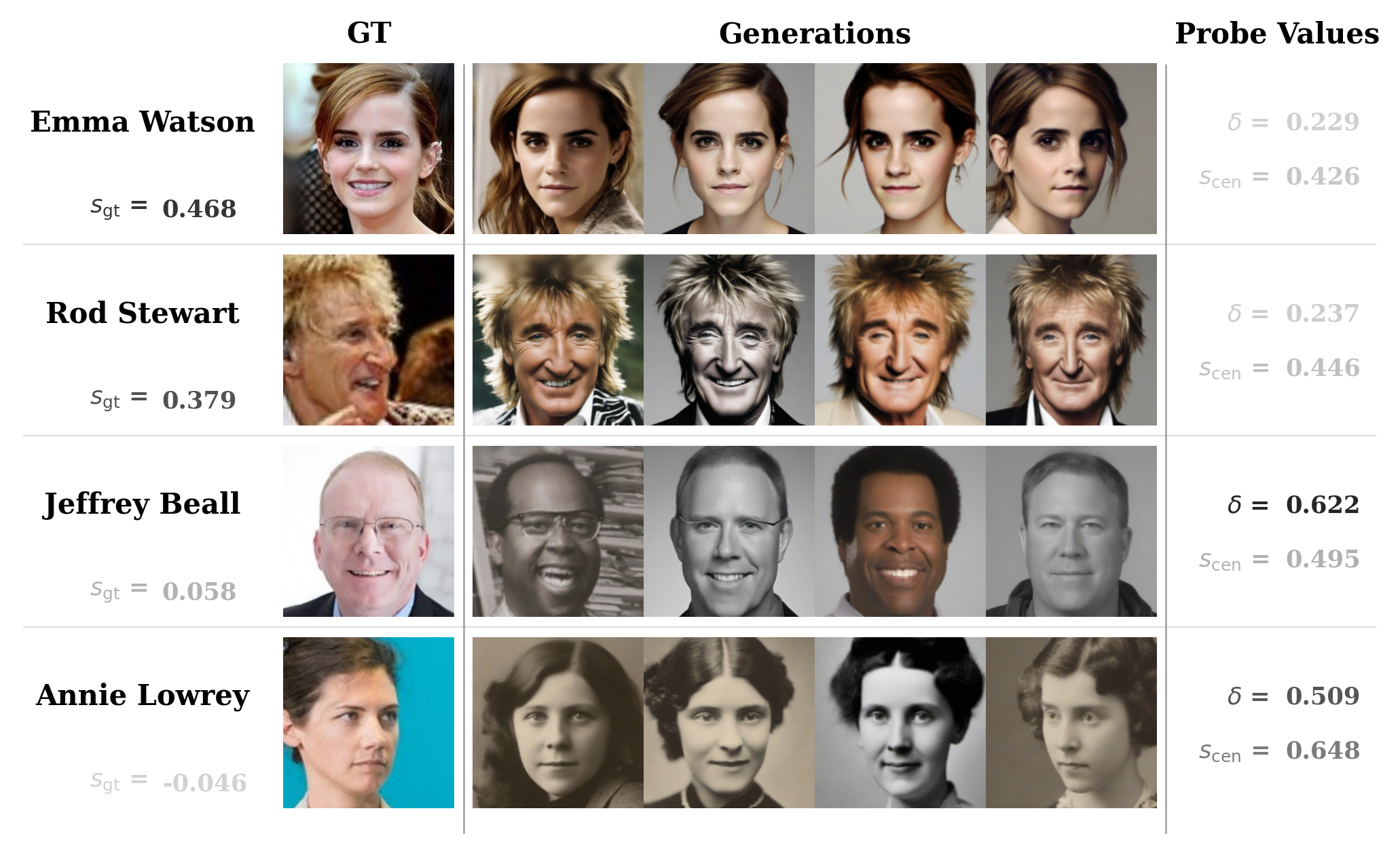}
  \caption{%
    Qualitative examples spanning the memorization spectrum for SDXL-Base.
    Each row shows a name's Wikipedia GT photo and four generated images, followed by probe values $\disp$ (dispersion) and $\censim{}$ (centroid similarity); the prediction target $\gtsim{}$ (reference similarity to GT) appears in the name label at left.
    All values are shaded darker with greater magnitude.
    Rows are ordered from high to low $\gtsim{}$: top rows show memorized identities whose generations resemble the GT and each other, while bottom rows show fabricated identities.
    Additional examples across all T2I models are shown in the appendix.
  }
  \label{fig:qual_gallery}
\end{figure*}

%% file: figures/qual_cross_model.tex
\begin{figure*}[t]
  \centering
  \includegraphics[width=\textwidth]{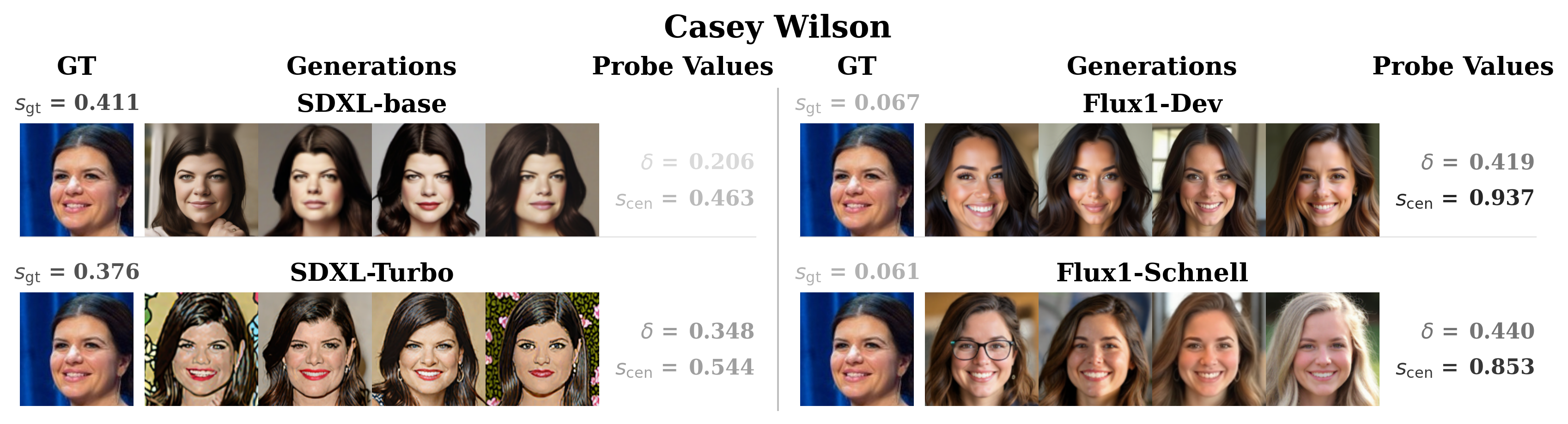}
  \caption{%
    Cross-model comparison for a single celebrity name (Casey Wilson).
    For each model, we show the GT face from \namesakes{}, generated faces and their reference similarity \gtsim{}, and our probe values (dispersion \disp{} and centroid similarity \censim{}). All values are shaded darker with greater magnitude.
    SDXL models (left) reproduce the identity more faithfully than Flux-based models (right)---reflected in reference similarity \gtsim{} values---and probe values inversely correlate with \gtsim{}, as expected.
  }
  \label{fig:qual_cross_model}
\end{figure*}

%% file: figures/demog_bar.tex
\begin{figure}[!ht]
    \centering
    \includegraphics[width=\linewidth]{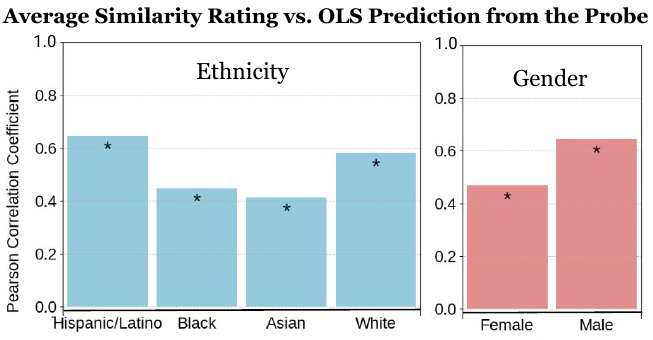}
    \caption{Comparison across single demographic attributes (Gender, Race) for the association between predicted memorization (OLS predictions on the probing scores) and the average human-rated similarity scores. $*$ indicates significance of the Pearson r, with $p < .05$.  }
    \label{fig:ols_vs_avgsim}
\end{figure}

%% file: sec/05_rw.tex
\section{Related Work}
\label{sec:related}

Our work is most related to studies on identity inference in vision-language models: methods identifying leakage of data related to an individual's identity, where we focus on data relating to names and facial imagery.
Prior works have explored this in the context of discriminative CLIP models~\citep{hintersdorf2024does,li2025tuni} and image generation models~\citep{webster2021person,vora2025identity}, but these works require either GT photos, access to training data, or white-box access to gradients.
By contrast, our probe targeting generative T2I models is fully black-box and reference-free.

A related line of work studies data memorization in image generation models such as diffusion models. Works on inference attacks~\citep{hu2023membership, duan2023diffusion, dubinski2024towards} study whether a given image was seen during training, and works on memorization~\citep{carlini2023extracting,somepalli2023diffusion,somepalli2023understanding} identify memorized images, but these do not address whether a holistic identity was memorized. \citet{wen2024detecting} find that memorized prompts steer generation more strongly towards seed-independent directions, supporting our dispersion probe.
However, their method requires access to intermediate 
model predictions during denoising and targets 
\emph{image-level} memorization (reproducing a specific 
training image), whereas our dispersion probe is fully 
black-box and targets \emph{identity-level} memorization 
(learning a coherent identity).

Our work also builds on research using behavioral signals to probe vision-language models for latent knowledge, including via input-level pseudoword
probes~\citep{alper2023kiki}, embedding geometry~\citep{alper2024emergent}, and character-level prompt perturbations~\citep{struppek2024exploiting}. \citet{yuan2023inserting} and \citet{zhao2025magicnaming} find that celebrity identities span a text embedding subspace; these support our geometric intuitions, although our behavioral probe does not depend on input embedding geometry.
\citet{luccioni2023stable} characterize the average face associated by T2I models to prompts, parallel to our visualization method using blended prototype faces (\Cref{sec:app_proto_blending}).

%% file: sec/06_conc.tex
\section{Conclusion}
\label{sec:conc}

In this work, we have introduced a black-box behavioral probe to distinguish between identity memorization and identity fabrication in T2I models applied to personal names. To benchmark performance, we have contributed the \namesakes{} dataset of identities stratified by fame, including names and ground-truth faces as well as perturbed names. Our results on \namesakes{} have shown that overall our probe can effectively diagnose identity memorization, revealing substantial differences between SOTA models in use today. Moreover, as this probe requires neither access to GT images nor model internals, it may generalize to any future T2I model agnostic of architecture, making it suitable for future privacy auditing.

We foresee various additional applications and extensions of our methodology: Future work could evaluate the effectiveness of identity-unlearning methods and whether memorization is unequally distributed among demographic groups. Another promising direction is mechanistic probing, for example, by finding interpretable circuits which encode identity recognition.

\section*{Limitations}
\label{sec:limitations}

A central limitation of our current methodology is the reliance on data and models which may under-represent diverse faces and identities. Our dataset reflects the English Wikipedia's demographic skew, with names and faces over-representing demographic groups such as English-speaking men.
This issue may also affect perturbed names, themselves drawn from English Wikipedia entries.
Future work could improve representation by collecting data from more diverse open sources, expanding beyond our limited \samplesize{} sample size, as well as grounding facial similarities in human annotations. In addition, facial similarity calculations use neural face embeddings, which have been documented to exhibit gender and racial bias~\citep{buolamwini2018gender,yucer2024racial}. 
As our framework is modular with respect to the face embedding model, future applications should use embedding models that work better for people of all skin tones and genders.

We have observed our probe to perform less strongly on distilled models, likely related to their known lack of diversity in generations. This is mitigated by an increased contribution of $\censim$ in our combined probe; as seen in \Cref{tab:models}, the combined probe still achieves meaningful performance ($R^2{=}0.33{-}0.44$ and $\text{AUC}
{>}0.77$).
Perturbed names fail face alignment somewhat more often than real names on SDXL models (\Cref{sec:app_face}); while exclusion is applied pairwise (removing the corresponding real name as well), this remains a mild limitation of the current benchmark methodology.
We also acknowledge that while our probe significantly predicts reference similarity (our continuous measure of identity memorization), this statistical relationship is not guaranteed to hold for a single sample, and our method should not be used as an exclusive diagnostic without human validation in performance-critical settings.
Finally, we do not account for public figures with identical names (e.g., ``Michael Jordan'' refers to both a famous basketball player and a well-known machine learning researcher), name changes, or those whose appearance has changed dramatically over time.

\section*{Ethical Considerations}
\label{sec:ethics}

The \namesakes{} dataset uses only openly-licensed images from Wikipedia, adhering to their respective licensing and attribution terms. While the individuals depicted did not explicitly consent to inclusion, we note that (1) all subjects are public figures according to Wikipedia's notability criteria, (2) the images were published under licenses permitting reuse, (3) the benchmark has protective intent for privacy auditing, such as flagging memorized identities in T2I models to inform unlearning or regulatory compliance, and (4) we include a datasheet~\citep{gebru2021datasheets} stipulating ethical use restrictions. We will also honor opt-out requests for removal from individuals included in the benchmark. 
Computer vision datasets embed disciplinary values and politics through curation choices~\citep{ stevens2021seeing}, like over-representing certain demographics; we explicitly report NAMESAKES's skew (Table~\ref{tab:demobreakdown}: majority White/men) as inherited from English Wikipedia's population. 
During benchmarking, generated images are used solely for calculating probe statistics and are not publicly released.

The facial recognition and T2I generation systems used in our pipeline are dual-use technologies. While we use them to evaluate identity recognition and privacy in existing models, the same existing tools can be used for surveillance, propagation of fake information, and stereotype perpetuation. 
Image databases often constructed with skewed race/gender categories, often amplifying biases in downstream facial analysis~\citep{scheuerman2021datasets} -- issues our probes inherit via embedding models~\citep{buolamwini2018gender, yucer2024racial}.
Use of these technologies must follow responsible guidelines for human-centered technologies, and we only condone use of our dataset and probe in this framework. In particular, as our probe may detect specific identities memorized by models, this information could in principle be maliciously exploited (e.g., by generating deepfake images of memorized identities); we recommend using this probe for audit purposes only, while highlighting its importance in flagging such privacy concerns. We do not recommend using our system as a conclusive determination of memorization for a given name and model in isolation, and encourage users to report demographic stratification in their results.
To address gaps in data traceability, our release also includes metadata on image sourcing and fame stratification, enabling data subjects or auditors to trace origins and request exclusions.

Our study involved manual annotation of image similarity using a crowdsourcing platform (Amazon Mechanical Turk). The task presented only images and annotation instructions; no personally identifiable information was collected or retained about the crowdworkers. Annotator identifiers were used only for task administration and quality control and were not included in the research dataset. We did not use any annotator attributes (e.g. demographics or individual‑level metadata) in our analysis, and no attempts were made to re‑identify individuals. Because the research uses de-identified annotation data and does not involve analysis of identifiable private information about living individuals, it qualifies as exempt from human subjects research oversight under our institutional policies.

%% file: sec/xx_app.tex
\section{Dataset Construction Details}
\label{sec:app_dataset}

\subsection{Data Source and Filtering}
\label{sec:app_names}

Our data is sourced from the official \href{https://dumps.wikimedia.org}{Wikimedia data dumps}, using data from the English Wikipedia. We use archived page contents from January 2026, and pageview counts from January 1, 2026. We select articles transcluding \texttt{Template:Infobox\_person}, using only pages with at least 10 views, yielding 189,697 pages. We use string heuristics to omit pages with malformed titles such as those containing punctuation, multi-person entries, and disambiguation suffixes (e.g., \emph{John Smith (actor)})). After stratified sampling over fame levels (\Cref{sec:dataset}) to obtain about 2K pages, we further filter by requiring a freely-licensed infobox image (CC*, Public Domain, or GFDL) with a successfully detected and aligned face, yielding the final \samplesize{} entries in \namesakes{}. Images are resized to have longest side length 256px and saved in PNG format.

\subsection{%
Construction of Perturbed Names}
\label{sec:app_doppel}

To construct perturbed %
names, we first create a pool of unique name components (such as first names and surnames) using the ${\sim}$190K names associated with selected Wikipedia infobox entries  (\Cref{sec:app_names}), split with word tokenization. Subsequently, in order to create a 
perturbed name for an existing name in \namesakes{}, we replace each name component with a candidate from the pool with minimum nonzero Levenshtein edit distance, subject to matching first letter. In cases of ties, one is selected at random. Random examples from \namesakes{} include the following: (shown in format \emph{real} $\to$ \emph{perturbed})
\begin{itemize}
\item \emph{Greg Abel} $\to$ \emph{Gren Adel}
\item \emph{Katie Holmes} $\to$ \emph{Kathie Holes}
\item \emph{J. Paul Gerry} $\to$ \emph{Jr. Poul Gatty}
\item \emph{Diana Ross} $\to$ \emph{Dyana Rossi}
\item \emph{Paul Whitehouse} $\to$ \emph{Paulo Whithouse}
\item \emph{Shreyas Talpade} $\to$ \emph{Shreyans Talmage}
\item \emph{James Nesbitt} $\to$ \emph{Jayes Nesbit}
\item \emph{Whitney Blake} $\to$ \emph{Whitley Blaže}
\item \emph{David Mickey Evans} $\to$ \emph{Davud Mickley Evins}
\item \emph{Ali Zafar} $\to$ \emph{Adi Zafer}
\end{itemize}

None of the perturbed %
names in \namesakes{} collide with real names from the ${\sim}$190K-name list of public figure entries with infoboxes and sufficient pageviews in Wikipedia. While this does not preclude that a perturbed %
name could match the name of a celebrity without a Wikipedia infobox, or that it could refer to a real person without a significant presence on Wikipedia, this confirms that these names are unlikely to correspond to public figures whose identities may have been memorized by T2I models trained on web-scale imagery. However, we leave open the possibility that additional identities may have been memorized by these models beyond the coverage of Wikipedia, which we foresee probes like ours detecting.

\section{Image Generation and Postprocessing}
\label{sec:app_generation}

\subsection{Computation Details}

All models are run on a single NVIDIA RTX A5000 GPU.

\input{tables/app_gen_config}

\subsection{Text-to-Image Settings}
\label{sec:app_models}

All images are generated using the fixed prompt template:
\begin{quote}
\texttt{``a picture of the person \{name\}''}
\end{quote}
where \texttt{\{name\}} is substituted with the personal name to be generated.

All T2I models are run at half or 8-bit precision, using checkpoints (from Hugging Face Hub) and settings listed in \Cref{tab:app_gen_config}. Generation settings use recommended defaults for compute-restricted settings (e.g., using the minimum number of recommended denoising steps). For SDXL-Base, the optional refiner module omitted.

For each name and model, we generate $k\!=\!4$ images in a single batch, with noise seeds fixed between generations. This value was chosen to balance between robustness and computational limitations; in particular, the larger T2I models tested would not be feasible to run on the entire \namesakes{} dataset within our time and computation budget if $k$ were significantly increased.

\subsection{Face Detection and Alignment}
\label{sec:app_face}

All generated images and GT photos are postprocessed via facial detection, alignment, and resizing. We use dlib's HOG-based face detector and landmark detector\footnote{\texttt{shape\_predictor\_68\_face\_landmarks.dat}} followed by warping with a similarity transformation into a canonical reference frame. The resized output is a $256{\times}256$\,px RGB image.

Edge cases where detection fails return a fully black image. This never occurs for GT images in \namesakes{}; for generated images, failure rates are low across models ($< 1.8\%$ of images). Rates are higher for 
perturbed names ($5.6\%-6.6\%$ of images for SDXL models; ${\leq}1.7\%$ for Flux models)---expected since fictional and obscure names are more likely to produce less face-like images. As black images would bias metric values (e.g., a name where all images are black would have zero dispersion, falsely signalling memorization) we apply a strict exclusion policy, excluding names from all analyses when any generated image fails face alignment. This leaves $n\!=\!1{,}215$--$1{,}224$ names for OLS and $n\!=\!996$--$1{,}186$ names for separability probes (for the latter, we exclude corresponding real and perturbed %
names if any generation for either of them fails face alignment).

\subsection{Face Embeddings}
\label{sec:app_facial_emb}

After undergoing alignment, we embed all facial imagery to calculate similarity scores, used for both reference similarity and probe calculations. For our embedding model, we use ArcFace~\cite{deng2019arcface} with the \texttt{w600k\_r50.onnx} model provided by InsightFace: a ResNet-50 backbone trained on WebFace600K (a.k.a. WebFace12M)~\citep{zhu2021webface260m}. This produces 512-dimensional embeddings which we L2-normalize before use.

\subsection{StyleGAN Inversion}
\label{sec:app_sg_inv}

For the face blending technique used to illustrate centroid similarity, we encode each of the $k$ generated faces (after alignment) with an E4E model~\citep{tov2021designing} into a latent code of shape $18 \times 512$. These $k$ latent codes are averaged elementwise and then passed through the frozen StyleGAN2 generator~\citep{Karras2019stylegan2} to produce the blended face image. We use frozen E4E and StyleGAN2 models that were trained on FFHQ~\citep{karras2019style}.

\section{Experimental Details}
\label{sec:app_stats}

Probes using linear regression probes fit an OLS regression model (\texttt{numpy.linalg.lstsq}) with an intercept term on the scores being used (\disp{} and \censim{} when using our full probe). Probes using logistic regression use defaults from scikit-learn (\texttt{sklearn.linear\_model.LogisticRegression}) and 1000 maximum iterations. These are fit and evaluated separately for each cross-validation fold.

For each cross-validation fold, centroid similarity \censim{} is computed inductively by only using names from the train split in calculations. In other words, \censim{} for a test item yields the similarity of the closest train set centroid.

\section{Computational Cost Analysis}
\label{sec:app_comp_cost}
$k{=}4$ generations per name requires only modest resources. Per-name probe computation takes seconds for fast models (e.g., SDXL-Turbo) to a few minutes for the largest (Flux1-Dev) on a single NVIDIA RTX A5000 GPU. Image generation is the main bottleneck; probe computation adds negligible overhead. 

\section{Additional Experiments and Results}

\subsection{Ablations}
\label{sec:app_emb_ablation}

\paragraph{Face Embedding.}
To ablate the effect of the chosen ArcFace face embedding model, we compare to results using our methodology using FaceNet~\citep{schroff2015facenet}--with an Inception-ResNet-V1 backbone pretrained on VGGFace2~\citep{cao2018vggface2}, loaded via the \texttt{facenet-pytorch} library. A comparison of results between these embedding models is provided in \Cref{tab:emb_ablation}. Overall performance is similar, though ArcFace is more performant, particularly for OLS probing with SDXL models, justifying our choice of ArcFace as our primary embedding model.

\input{tables/embedding_ablation}

\paragraph{Number of generations ($k$).}
Subsampling existing generations to $k{=}2$ yields 
$R^2$ of $0.49$ vs.\ $0.58$ ($k{=}4$) for SDXL-Base 
and $0.25$ vs.\ $0.35$ for Flux1-Dev. Separability 
(AUC) is more consistent, dropping only $0.01$--$0.05$ 
across models. $k{=}4$ thus provides a meaningful 
improvement while remaining feasible with our computational resources.

\paragraph{Reference pool size for $\censim$.}
Restricting the pool of names used to calculate $\censim$ to $500$ and $250$ names 
($3$ random subsamples each), $R^2$ drops $\leq 0.03$ 
and AUC drops $\leq 0.01$ across all models, 
demonstrating robustness to this pool size.

\paragraph{Prompt template.}
To assess sensitivity to prompt wording, we re-run generation and probing on SDXL-Turbo with two alternative templates. \texttt{``portrait of the person \{name\}''} gives results consistent with our main template ($R^2{=}0.444$ vs. $0.438$; $\text{AUC}{=}0.800$ vs. $0.772$), while the bare \texttt{``the person \{name\}''} yields a less performant probe ($R^2{=}0.345$; $\text{AUC}{=}0.741$), coinciding with higher exclusion due to alignment failure (13\% of names vs. 3–4\%). This reflects a less suitable prompt for generating the facial images needed for face-based probing.

\subsection{Fame-Stratified Separability}
\label{sec:app_fame_sep}

\Cref{tab:fame_sep} reports real-vs.-perturbed %
name separability separately for
high- and low-fame names, defined as those with fame (log-pageview) scores above or below the median. Calculations use cross-validation following our main results. High-fame names show substantially higher separability across models, expected since both low-fame and %
perturbed names are unlikely to be memorized.
Low-fame names exhibit low separability as expected, since these names are unlikely to have been memorized; the probe correctly assigns them scores comparable to their perturbations.
This stratification supports the intended interpretation of our method: the probe identifies memorization where it occurs and reports its absence elsewhere, rather than producing spurious positives for unfamiliar names.

\input{tables/fame_separability}

\input{figures/qual_gallery_app}

\subsection{Additional Metrics}

\input{tables/app_prf1}

For the real-vs.-perturbed separability task (all names, threshold 0.5), precision/recall/F1 across CV folds are provided in \Cref{tab:prf1}; these are seen to be consistent with the AUC and accuracy values in \Cref{tab:models} (e.g., best results for SDXL-Base).

\subsection{Demographic Disaggregation}
\label{sec:app_demog_disagg}

\Cref{tab:disaggregated_analysis,tab:auc_disaggregated} disaggregate our two main evaluation metrics, $R^2$ (predicting reference similarity) and AUC (real-vs.-perturbed separability), by our race and gender annotations (\Cref{sec:dataset}). Disaggregated $R^2$ is lower for faces labeled Black or woman across most models, consistent with known racial and gender biases in face embedding models~\citep{buolamwini2018gender,yucer2024racial}, which underlie both our probe and our reference-similarity target. However, disaggregated AUC does not follow the same pattern: separability is highest for Black names on SDXL-Base (0.96) and SDXL-Turbo (0.85), and comparable or higher for names of women versus men across most models.
We do not have a full explanation for this divergence. One contributing factor may be that $R^2$ and AUC are differently sensitive to bias in the reference-similarity target itself, since reference similarity is computed using the same face embedding model as our probes: if embeddings are systematically less accurate for some demographic groups, this adds noise to the regression target independently of probe quality, in a way that need not affect binary separability in the same direction. A second possible factor is that perturbed names are constructed by orthographic similarity to real names (\Cref{sec:app_doppel}) without controlling for demographic association, so real-vs.-perturbed separability may partly reflect incidental demographic differences between paired names rather than probe quality alone. We leave disentangling these factors, and the broader question of demographic performance disparities in identity-memorization probing, to future work.
\input{tables/app_r2_disagg}
\input{tables/app_auc_disagg}

\subsection{Additional Qualitative Results}
\label{sec:app_exemplars}

\Cref{fig:qual_gallery_app_sdxl,fig:qual_gallery_app_sdxl_turbo,fig:qual_gallery_app_flux_dev,fig:qual_gallery_app_flux_schnell}
show qualitative results in the format of
\Cref{fig:qual_gallery} for all T2I models under consideration. Comparing across figures illustrates the model differences
discussed in \Cref{sec:exp}: SDXL models exhibit more identity memorization than Flux models,
with SDXL-Base showing the strongest effect.

\input{figures/interp}

\subsection{Prototype Blending}
\label{sec:app_proto_blending}

We introduce an interpretability
technique, \emph{prototype blending}, that visualizes overall visual associations with unfamiliar names (\Cref{fig:interp}). For each name, we invert a set of generated faces into the latent space of a pretrained
StyleGAN2~\citep{Karras2019stylegan2} using an E4E
encoder model~\citep{tov2021designing} and decode their mean via the StyleGAN2 generator to produce a
\emph{prototype face} image illustrating the model's overall
association for that name. As seen in the figure, prototypes for unknown names reflect aggregate demographic associations such as gender, race, and age, providing a visual tool for understanding T2I models' stereotypical associations with names.

\section{Human Evaluation Details}                                                                                                
  \label{sec:survey-interface}           

  \input{figures/survey_instructions}

\subsection{Platform and Participants}

For our human evaluation, we recruit participants on Amazon Mechanical Turk, without geographic limitations.

Each participant received \$0.50 in compensation for a single HIT (approximately 3 minutes in duration), which covered 25 names (one-fourth of the set under consideration). We collected a total of 38 HITs across four such splits, targeting at least 9 responses per name.

Our survey involved collecting image similarity annotations from crowdworkers, without collecting personally identifiable information, sensitive data, or research questions regarding annotators themselves. 

\subsection{Survey Methodology}

Each survey item follows a two-step design: annotators first indicate whether they recognize the person by name (shown alongside a single GT photo), then rate the similarity between the GT photo and a generated image on a 5-point Mean Opinion Score (MOS) scale (1\,=\,Different person, 5\,=\,Same person). Each survey includes a hidden control pair (two real photos of the same person); annotators rating the control below 4 are excluded---in total, excluding two out of 38 HITs. After filtering, we retain 9 valid annotators per item (items were split between multiple HITs).

\subsection{Survey Interface}
  
\Cref{fig:survey-interface} shows the annotation interface used in our human evaluation study. Each annotator is first presented    
with task instructions and the rating scale (\cref{fig:survey-instructions}). They then evaluate each celebrity in a two-step card  
layout (\cref{fig:survey-card}): first indicating whether they recognize the person, then rating the similarity between a           
ground-truth photo and a generated image on a 1--5 Mean Opinion Score (MOS) scale.

\section{Generative AI Disclosure}

Generative AI tools were used for polishing wording and formatting in this manuscript, and for coding assistance. Gemini was used for demographic annotation as discussed in \Cref{sec:dataset}; these annotations are not included in the dataset for release.

%% file: tables/app_gen_config.tex
\begin{table*}[h]
  \centering
  \small
  \begin{tabular}{lcccccc}
    \toprule
    Model & Checkpoint & \#Params & Precision & Steps & CFG & Resolution \\
    \midrule
    SDXL-Base    & \texttt{sd\_xl\_base\_1.0}       & 3.5B & fp16 & 20 & 7.0  & $1024 \times 1024$ \\
    SDXL-Turbo   & \texttt{sd\_xl\_turbo\_1.0\_fp16} & 3.5B & fp16 & 4  & 1.0  & $512 \times 512$  \\
    Flux1-Dev    & \texttt{flux1-dev-fp8}            & 12B  & fp8  & 20 & 3.5  & $1024 \times 1024$ \\
    Flux1-Schnell& \texttt{flux1-schnell}            & 12B  & bf16 & 1  & ---  & $1024 \times 1024$ \\
    \bottomrule
  \end{tabular}
  \caption{Generation settings per model.
    Flux1-Schnell uses timestep distillation; CFG is not applicable).}
  \label{tab:app_gen_config}
\end{table*}

%% file: tables/embedding_ablation.tex
\begin{table*}[t]
  \centering
  \small
  \begin{tabular}{lccccc}
  \toprule
  & \multicolumn{3}{c}{Predicting Reference Similarity ($R^2$)} & \multicolumn{2}{c}{Real vs.\ Perturbed%
  } \\
  \cmidrule(lr){2-4} \cmidrule(lr){5-6}
  Model & \disp{} only & \censim{} only & Both & AUC & Acc \\
  \midrule
  \multicolumn{6}{l}{\textit{ArcFace (primary)}} \\
    SDXL-Base & 0.547 {\scriptsize $\pm$0.031} & 0.128 {\scriptsize $\pm$0.030} & 0.581 {\scriptsize $\pm$0.040} & 0.859 {\scriptsize $\pm$0.015} & 0.791 {\scriptsize $\pm$0.014} \\
    SDXL-Turbo & 0.288 {\scriptsize $\pm$0.050} & 0.171 {\scriptsize $\pm$0.052} & 0.438 {\scriptsize $\pm$0.044} & 0.772 {\scriptsize $\pm$0.036} & 0.723 {\scriptsize $\pm$0.029} \\
    Flux1-dev & 0.218 {\scriptsize $\pm$0.054} & 0.150 {\scriptsize $\pm$0.007} & 0.349 {\scriptsize $\pm$0.060} & 0.781 {\scriptsize $\pm$0.033} & 0.750 {\scriptsize $\pm$0.018} \\
    Flux1-schnell & 0.137 {\scriptsize $\pm$0.078} & 0.187 {\scriptsize $\pm$0.036} & 0.325 {\scriptsize $\pm$0.056} & 0.785 {\scriptsize $\pm$0.016} & 0.748 {\scriptsize $\pm$0.011} \\
  \midrule
  \multicolumn{6}{l}{\textit{FaceNet}} \\
    SDXL-Base & 0.478 {\scriptsize $\pm$0.012} & 0.043 {\scriptsize $\pm$0.029} & 0.492 {\scriptsize $\pm$0.020} & 0.846 {\scriptsize $\pm$0.011} & 0.769 {\scriptsize $\pm$0.019} \\
    SDXL-Turbo & 0.208 {\scriptsize $\pm$0.070} & 0.059 {\scriptsize $\pm$0.017} & 0.288 {\scriptsize $\pm$0.057} & 0.777 {\scriptsize $\pm$0.030} & 0.727 {\scriptsize $\pm$0.028} \\
    Flux1-dev & 0.275 {\scriptsize $\pm$0.059} & 0.091 {\scriptsize $\pm$0.035} & 0.331 {\scriptsize $\pm$0.067} & 0.784 {\scriptsize $\pm$0.028} & 0.734 {\scriptsize $\pm$0.032} \\
    Flux1-schnell & 0.252 {\scriptsize $\pm$0.039} & 0.064 {\scriptsize $\pm$0.019} & 0.309 {\scriptsize $\pm$0.044} & 0.809 {\scriptsize $\pm$0.004} & 0.733 {\scriptsize $\pm$0.008} \\
  \bottomrule
  \end{tabular}
  \caption{
    Embedding model ablation: ArcFace vs.\ FaceNet.
    All columns as in \Cref{tab:models}.
    Results are comparable between models, with ArcFace generally outperforming FaceNet, justifying our choice of ArcFace for our main results.
  }
  \label{tab:emb_ablation}
\end{table*}

%% file: tables/fame_separability.tex
\begin{table*}[t]
  \centering
  \small
  \begin{tabular}{lcccc}
  \toprule
   & \multicolumn{2}{c}{AUC} & \multicolumn{2}{c}{Acc} \\
   \cmidrule(lr){2-3} \cmidrule(lr){4-5}
  Model & High Fame & Low Fame & High Fame & Low Fame \\
  \midrule
    SDXL-Base     & 0.947 {\scriptsize $\pm$0.011} & 0.760 {\scriptsize $\pm$0.032} & 0.902 {\scriptsize $\pm$0.014} & 0.697 {\scriptsize $\pm$0.040} \\
    SDXL-Turbo    & 0.835 {\scriptsize $\pm$0.024} & 0.650 {\scriptsize $\pm$0.047} & 0.777 {\scriptsize $\pm$0.019} & 0.606 {\scriptsize $\pm$0.044} \\
    Flux1-Dev     & 0.888 {\scriptsize $\pm$0.021} & 0.639 {\scriptsize $\pm$0.048} & 0.832 {\scriptsize $\pm$0.021} & 0.629 {\scriptsize $\pm$0.039} \\
    Flux1-Schnell & 0.844 {\scriptsize $\pm$0.018} & 0.681 {\scriptsize $\pm$0.043} & 0.792 {\scriptsize $\pm$0.015} & 0.651 {\scriptsize $\pm$0.039} \\
  \bottomrule
  \end{tabular}
  \caption{
    Fame-stratified separability (real vs. perturbed%
    ):
    High-fame names (log-pageviews above the median) are more separable than low-fame names,
    consistent with stronger identity memorization for well-known individuals.
  }
  \label{tab:fame_sep}
\end{table*}

%% file: figures/qual_gallery_app.tex
\begin{figure*}[p]
  \centering
  \includegraphics[width=\textwidth]{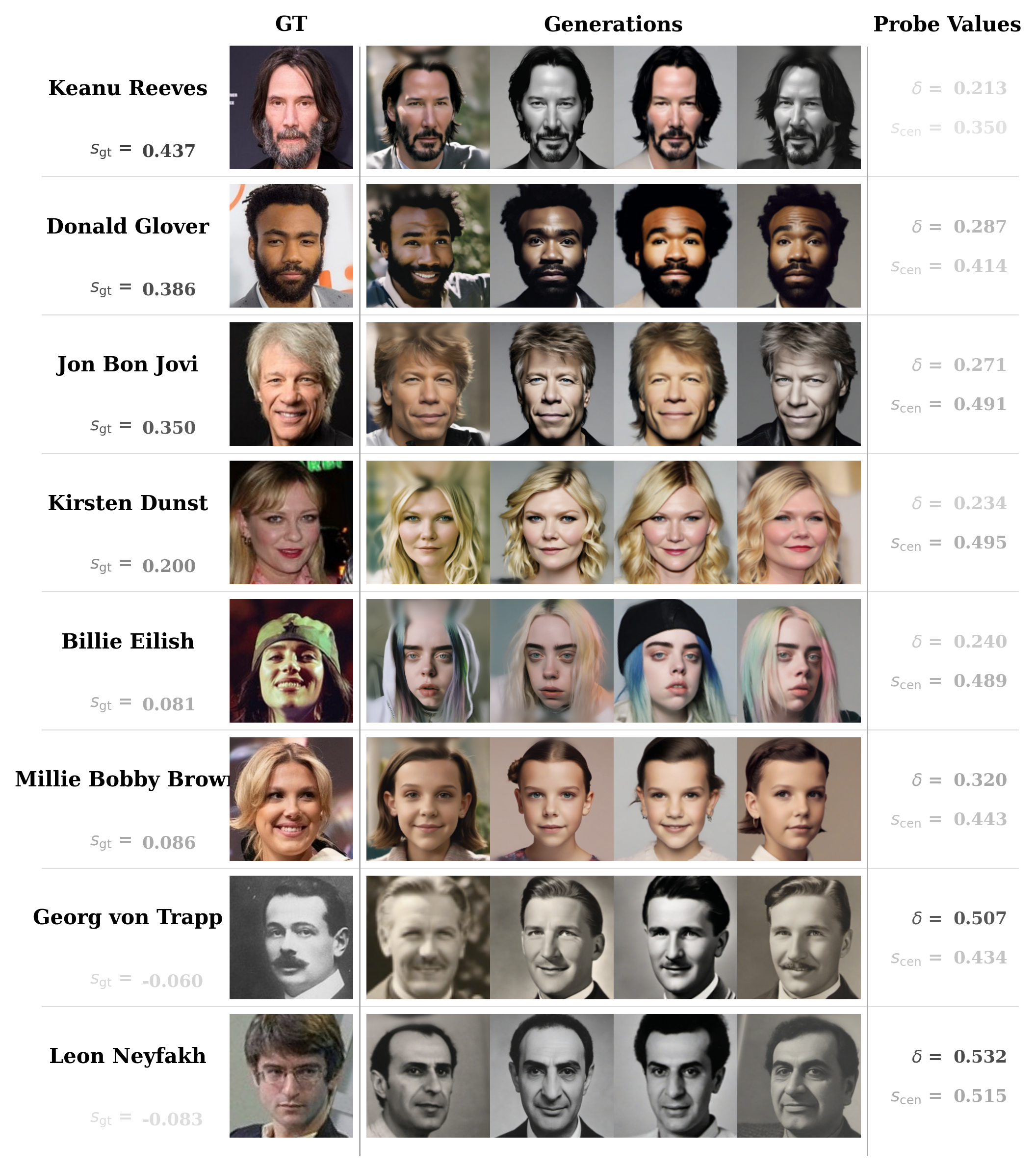}
  \caption{%
    Additional qualitative examples (SDXL-Base) spanning the memorization spectrum.
    Layout as in \Cref{fig:qual_gallery}.
    Values are shaded darker with greater magnitude.%
  }
  \label{fig:qual_gallery_app_sdxl}
\end{figure*}

\begin{figure*}[p]
  \centering
  \includegraphics[width=\textwidth]{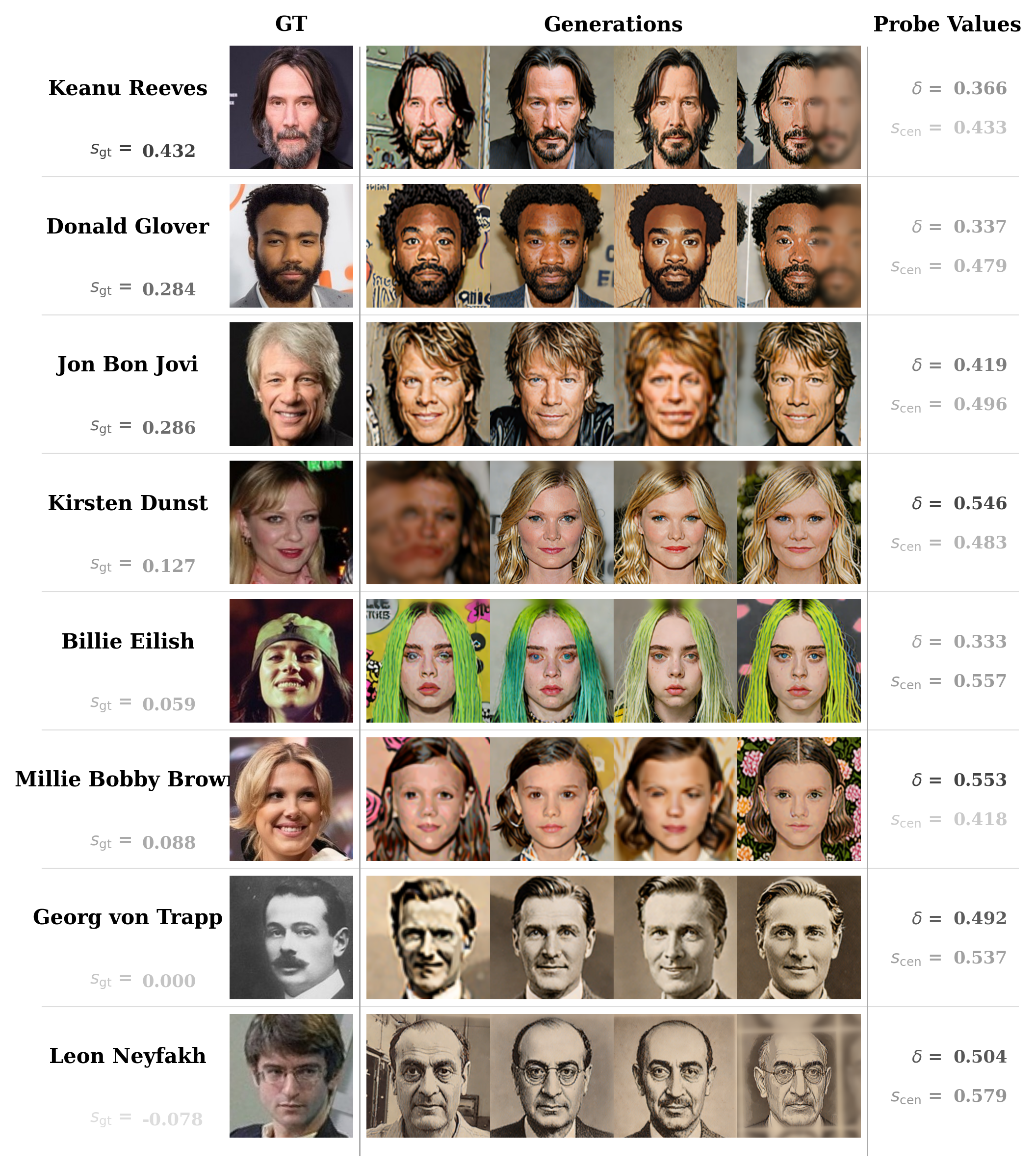}
  \caption{%
    Additional qualitative examples (SDXL-Turbo). Same names and layout as
    \Cref{fig:qual_gallery_app_sdxl}.%
  }
  \label{fig:qual_gallery_app_sdxl_turbo}
\end{figure*}

\begin{figure*}[p]
  \centering
  \includegraphics[width=\textwidth]{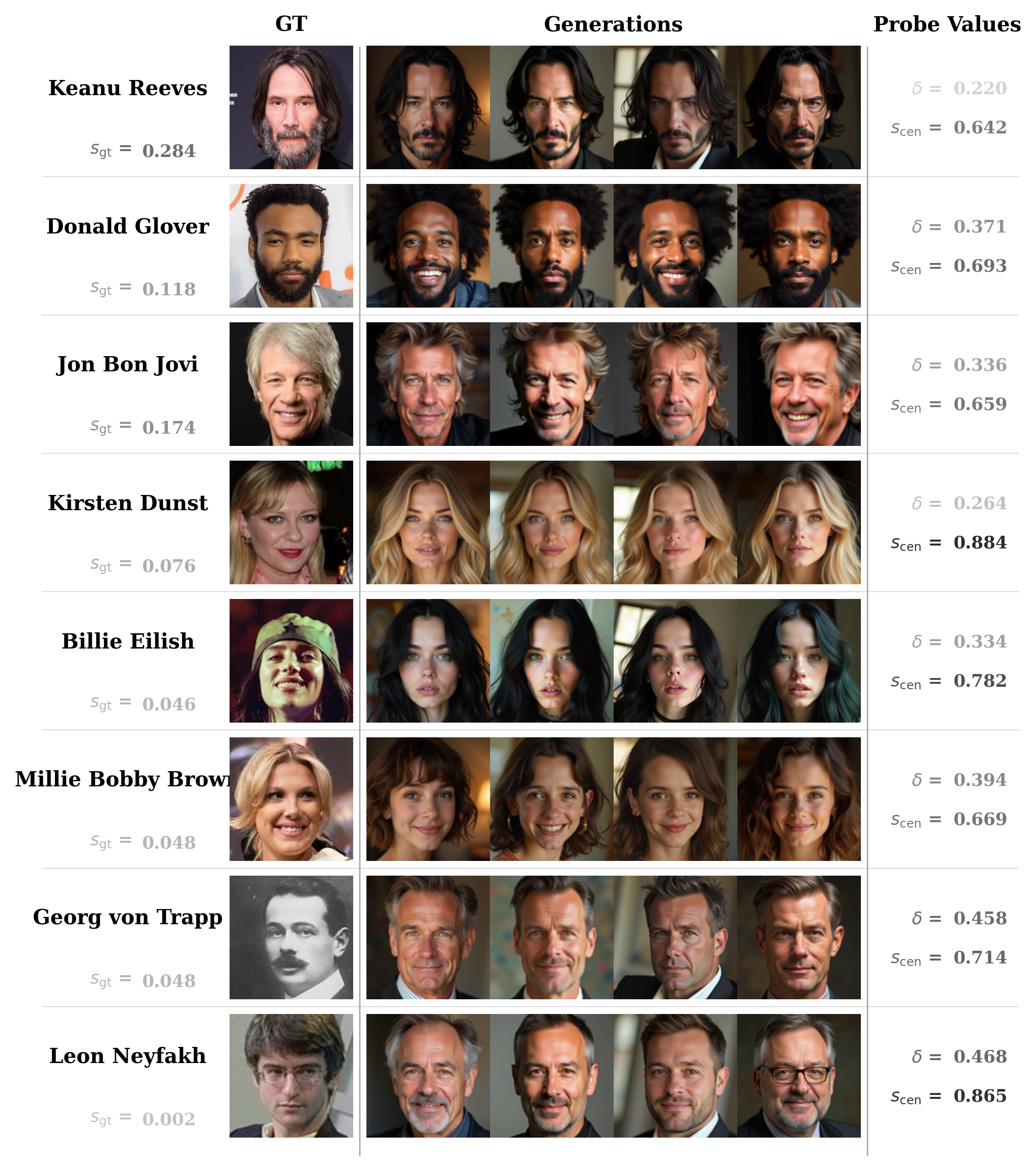}
  \caption{%
    Additional qualitative examples (Flux1-Dev). Same names and layout as
    \Cref{fig:qual_gallery_app_sdxl}.%
  }
  \label{fig:qual_gallery_app_flux_dev}
\end{figure*}

\begin{figure*}[p]
  \centering
  \includegraphics[width=\textwidth]{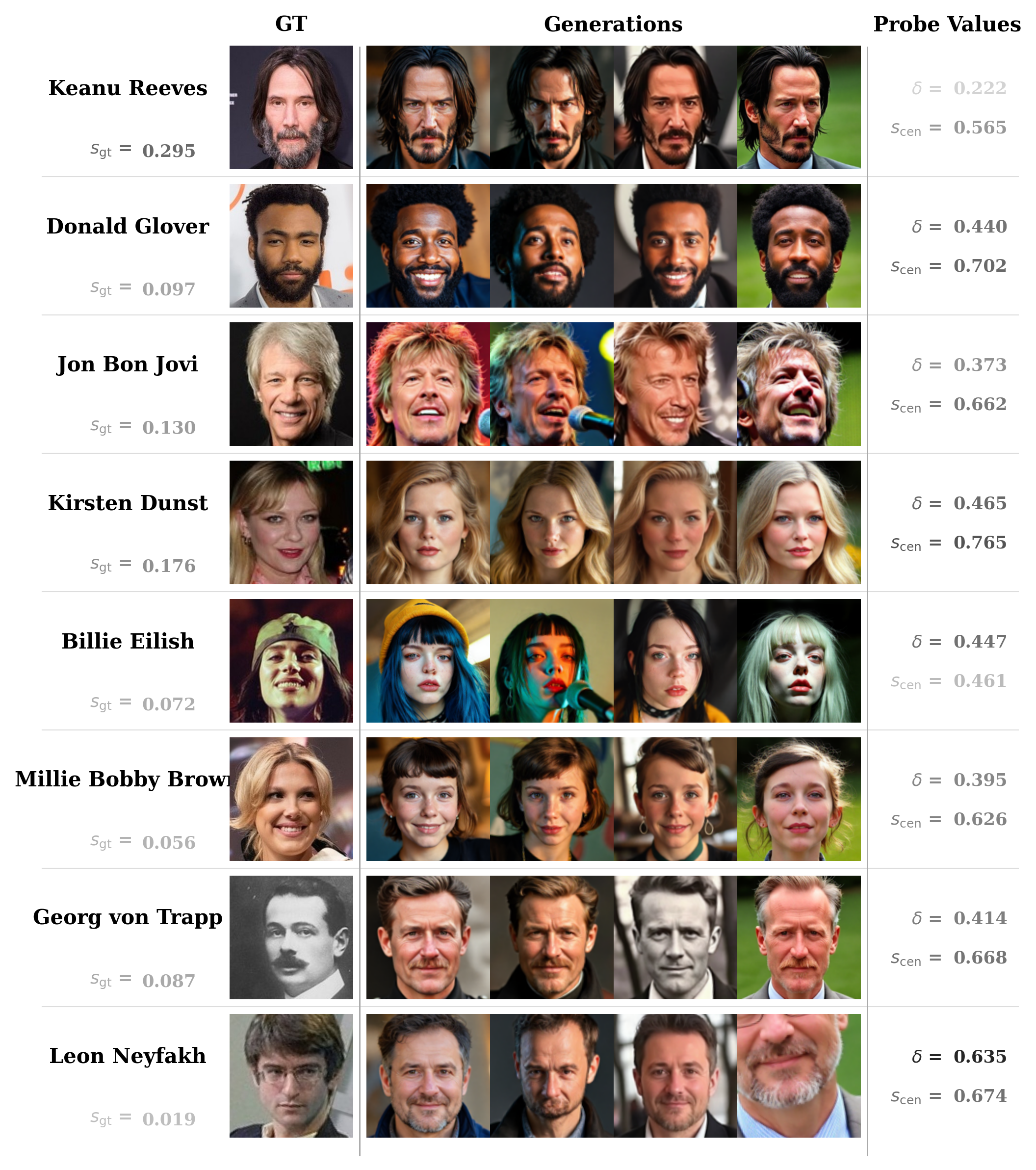}
  \caption{%
    Additional qualitative examples (Flux1-Schnell). Same names and layout as
    \Cref{fig:qual_gallery_app_sdxl}.%
  }
  \label{fig:qual_gallery_app_flux_schnell}
\end{figure*}

%% file: tables/app_prf1.tex
\begin{table}[h]
\centering
\small
\begin{tabular}{lccc}
\toprule
Model & Precision & Recall & F1 \\
\midrule
SDXL-Base     & 0.832 \tiny{$\pm$0.011} & 0.728 \tiny{$\pm$0.025} & 0.776 \tiny{$\pm$0.017} \\
SDXL-Turbo    & 0.744 \tiny{$\pm$0.030} & 0.681 \tiny{$\pm$0.034} & 0.711 \tiny{$\pm$0.031} \\
Flux1-Dev     & 0.784 \tiny{$\pm$0.016} & 0.691 \tiny{$\pm$0.061} & 0.733 \tiny{$\pm$0.033} \\
Flux1-Schnell & 0.780 \tiny{$\pm$0.024} & 0.693 \tiny{$\pm$0.036} & 0.733 \tiny{$\pm$0.015} \\
\bottomrule
\end{tabular}
\caption{Additional classification metrics for real vs.\ perturbed name separability (all names, threshold 0.5), complementing the AUC and accuracy values in \Cref{tab:models}. Mean $\pm$ std across CV folds.}
\label{tab:prf1}
\end{table}

%% file: tables/app_r2_disagg.tex
\begin{table}[ht]
\centering
\resizebox{\linewidth}{!}{
\begin{tabular}{lcccc|cc}
\toprule
 & \multicolumn{4}{c}{Race} & \multicolumn{2}{c}{Gender} \\
\cline{2-5} \cline{6-7}
Model & Hispanic/Latino & Asian & White &  Black  & Men & Women \\
\midrule
SDXL-Base & 0.66 {\scriptsize (n=44)} & 0.54 {\scriptsize (n=157)} & 0.60 {\scriptsize (n=901)} & 0.39 {\scriptsize (n=65)} & 0.59 {\scriptsize (n=712)} & 0.58 {\scriptsize (n=502)} \\
SDXL-Turbo & 0.39 {\scriptsize (n=41)} & 0.38 {\scriptsize (n=160)} & 0.46 {\scriptsize (n=904)} & 0.40 {\scriptsize (n=65)} & 0.50 {\scriptsize (n=713)} & 0.37 {\scriptsize (n=505)} \\
Flux1-Dev & 0.34 {\scriptsize (n=43)} & 0.26 {\scriptsize (n=156)} & 0.36 {\scriptsize (n=909)} & 0.33 {\scriptsize (n=68)} & 0.44 {\scriptsize (n=717)} & 0.15 {\scriptsize (n=507)} \\
Flux1-Schnell & 0.39 {\scriptsize (n=42)} & 0.30 {\scriptsize (n=161)} & 0.35 {\scriptsize (n=899)} & 0.28 {\scriptsize (n=64)} & 0.39 {\scriptsize (n=712)} & 0.25 {\scriptsize (n=502)} \\
\bottomrule
\end{tabular}
}
\caption{
$R^2$ predicting reference similarity using cross-validation, disaggregated by race and gender, with sample sizes provided. These are the same unit as the ``Both'' column in the left half of \Cref{tab:models}.
}
\label{tab:disaggregated_analysis}
\end{table}

%% file: tables/app_auc_disagg.tex
\begin{table}[ht]
\centering
\resizebox{\linewidth}{!}{
\begin{tabular}{lcccc|cc}
\toprule
 & \multicolumn{4}{c}{Race} & \multicolumn{2}{c}{Gender} \\
\cline{2-5} \cline{6-7}
Model & Hispanic/Latino & Asian & White &  Black  & Men & Women \\
\midrule
SDXL-Base & 0.85 {\scriptsize $\pm$0.10 (n=68)} & 0.77 {\scriptsize $\pm$0.04 (n=284)} & 0.86 {\scriptsize $\pm$0.02 (n=1462)} & 0.96 {\scriptsize $\pm$0.04 (n=106)} & 0.84 {\scriptsize $\pm$0.02 (n=1130)} & 0.88 {\scriptsize $\pm$0.02 (n=860)} \\
SDXL-Turbo & 0.69 {\scriptsize $\pm$0.18 (n=70)} & 0.59 {\scriptsize $\pm$0.04 (n=298)} & 0.79 {\scriptsize $\pm$0.04 (n=1636)} & 0.85 {\scriptsize $\pm$0.09 (n=114)} & 0.77 {\scriptsize $\pm$0.03 (n=1286)} & 0.78 {\scriptsize $\pm$0.05 (n=918)} \\
Flux1-Dev & 0.77 {\scriptsize $\pm$0.17 (n=82)} & 0.70 {\scriptsize $\pm$0.04 (n=302)} & 0.79 {\scriptsize $\pm$0.04 (n=1768)} & 0.74 {\scriptsize $\pm$0.11 (n=128)} & 0.78 {\scriptsize $\pm$0.04 (n=1380)} & 0.79 {\scriptsize $\pm$0.03 (n=992)} \\
Flux1-Schnell & 0.66 {\scriptsize $\pm$0.26 (n=82)} & 0.69 {\scriptsize $\pm$0.07 (n=310)} & 0.80 {\scriptsize $\pm$0.03 (n=1738)} & 0.79 {\scriptsize $\pm$0.04 (n=126)} & 0.79 {\scriptsize $\pm$0.03 (n=1380)} & 0.79 {\scriptsize $\pm$0.04 (n=970)} \\
\bottomrule
\end{tabular}
}
\caption{
AUC for real-vs.-perturbed name separability using cross-validation, disaggregated by race and gender (mean\,$\pm$\,std across folds), with sample sizes provided.
}
\label{tab:auc_disaggregated}
\end{table}

%% file: figures/interp.tex
\begin{figure*}[t]
  \centering
  \setlength{\tabcolsep}{4pt}
  \begin{tabular}{ccc}
    \includegraphics[width=0.31\textwidth]{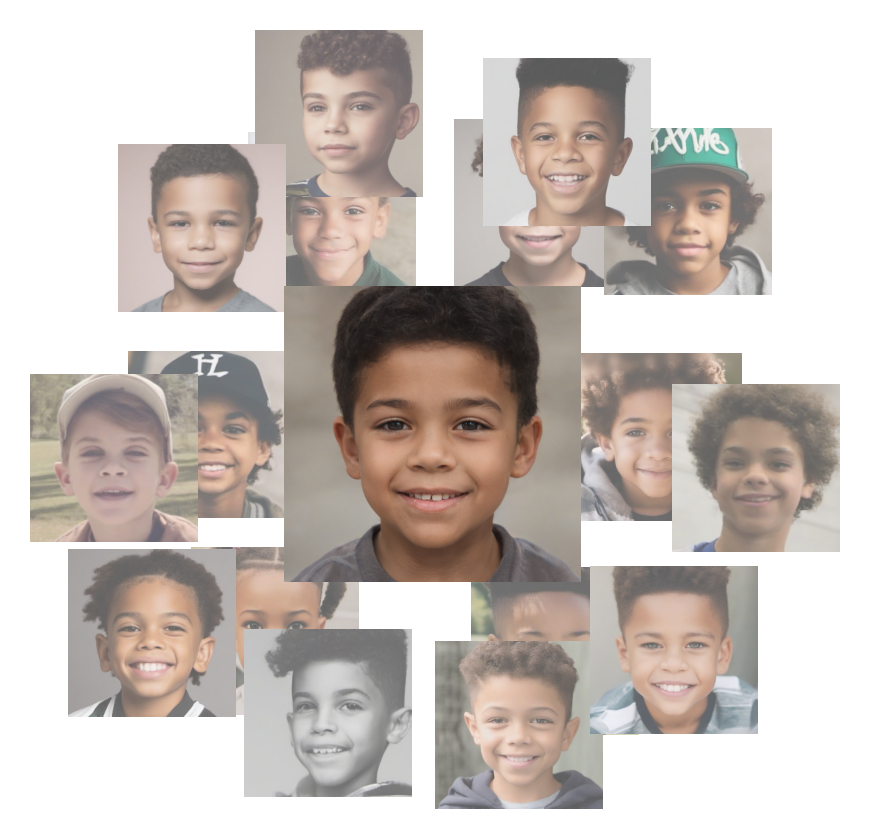} &
    \includegraphics[width=0.31\textwidth]{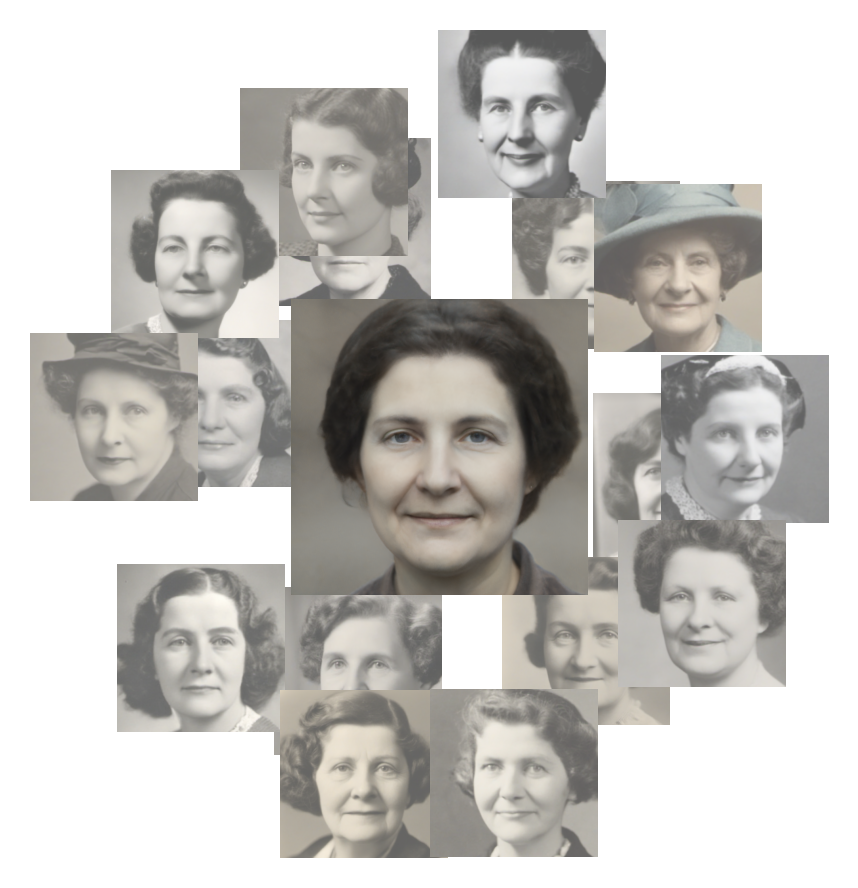} &
    \includegraphics[width=0.31\textwidth]{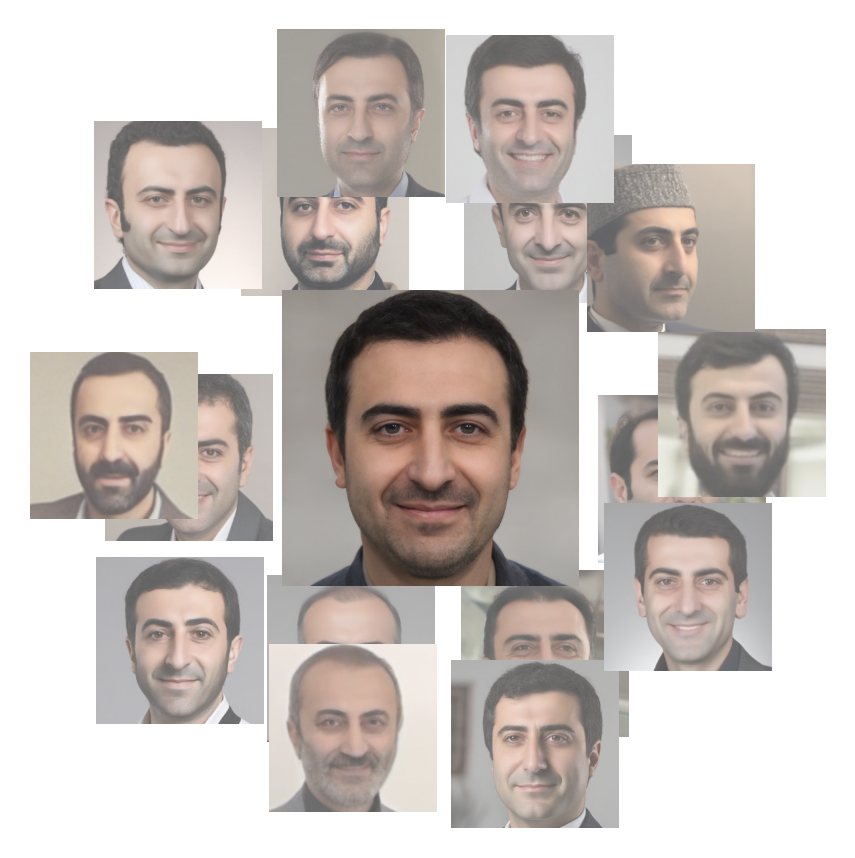} \\[-4pt]
    \textbf{Zayden Chuckson}
    & \textbf{Gertrude Schwartz}
    & \textbf{Mehmet Demir} \\
  \end{tabular}
  \caption{%
    \emph{Prototype blending}, an interpretability technique for visualizing associations with unfamiliar names.
    For each name, generated faces (faded, periphery) are inverted into StyleGAN2 latent vectors, and their mean code is decoded to produce a \emph{prototype face} (center,
    full color).
    Despite diverse generations, this yields prototype images that reflect their shared aggregate characteristics, such as demographic associations (e.g., gender, race, and age).
     Note that this technique is not used as part of our probe; it is a post-hoc interpretability visualization. All generations use SDXL-Base.
  }
  \label{fig:interp}
\end{figure*}

%% file: figures/survey_instructions.tex
\begin{figure*}[t]                                                                                                                   
    \centering                                              
    \begin{subfigure}[t]{0.48\linewidth}                                                                                              
      \centering                                            
      \includegraphics[width=\linewidth]{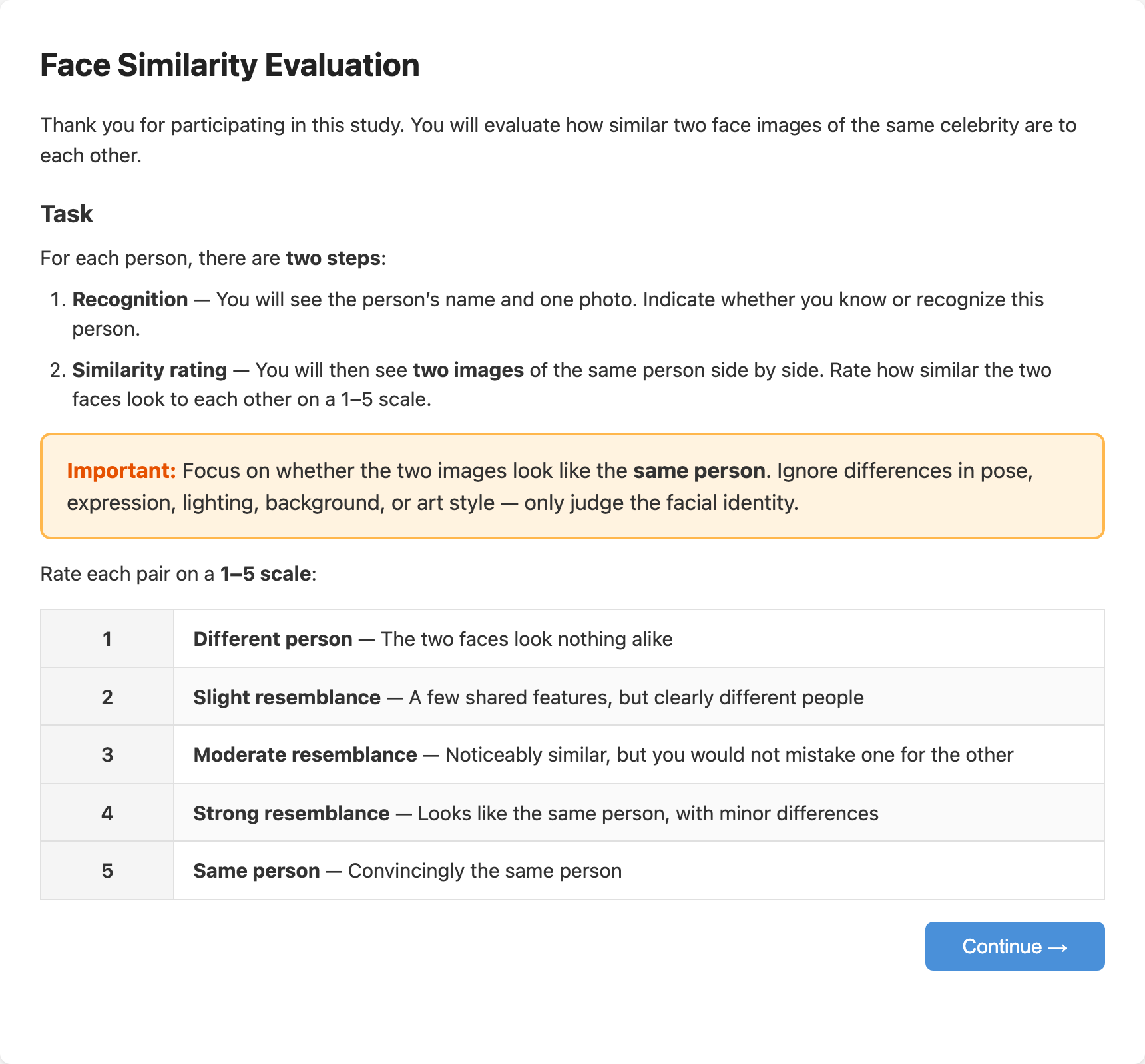}                                                                    
      \caption{Task instructions shown to annotators.}                                                                                
      \label{fig:survey-instructions}
    \end{subfigure}
    \hfill
    \begin{subfigure}[t]{0.48\linewidth}
      \centering
      \includegraphics[width=\linewidth]{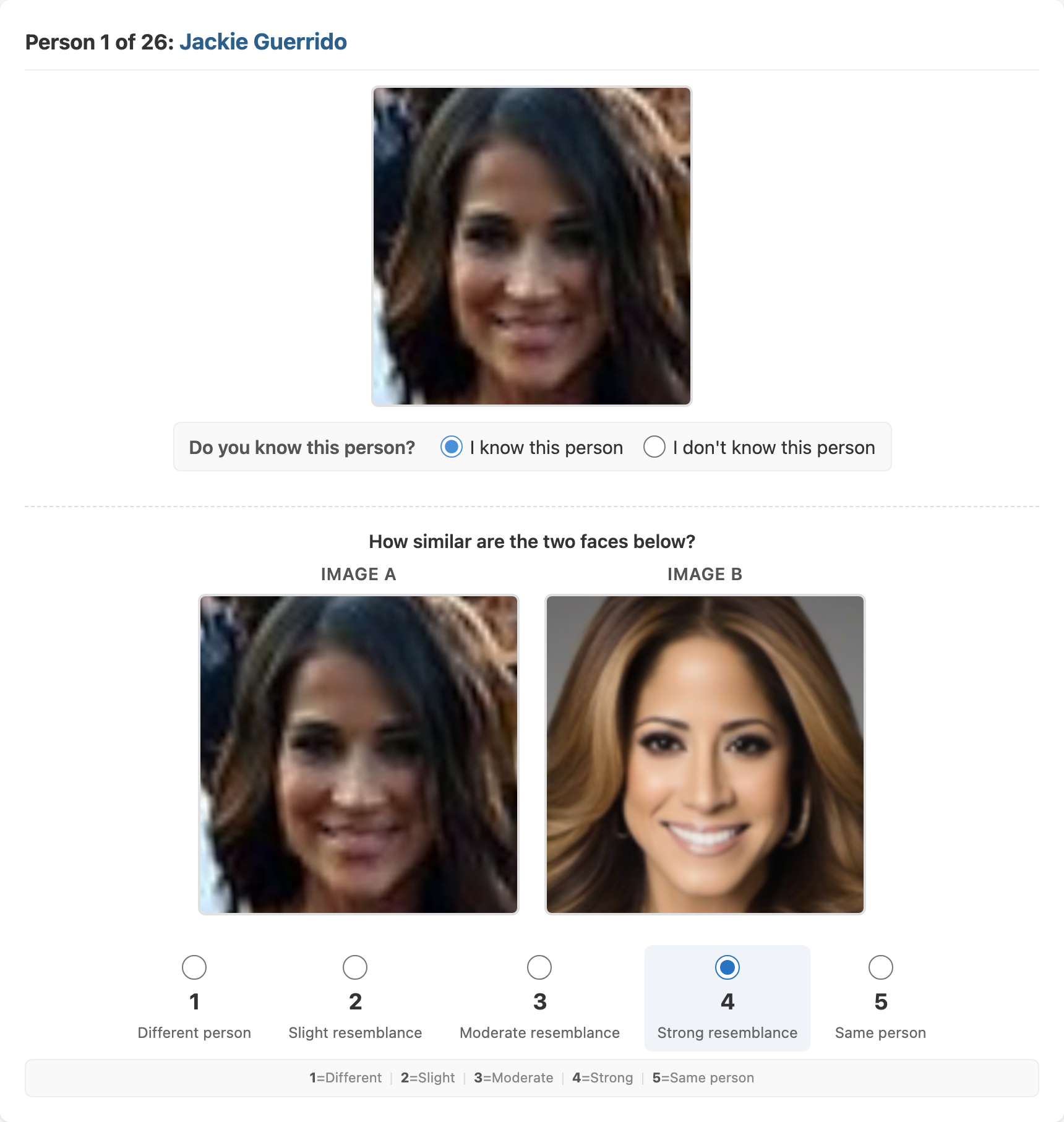}
      \caption{Evaluation card: recognition question followed by side-by-side similarity rating.}
      \label{fig:survey-card}
    \end{subfigure}
    \caption{User study interface. (a)~Annotators first read the task guidelines and rating scale. (b)~For each celebrity, they
  indicate whether they recognize the person, then rate the similarity between two images on a 1--5 scale.}
    \label{fig:survey-interface}
  \end{figure*}